\documentclass{article}

\PassOptionsToPackage{numbers, compress}{natbib}

 \usepackage[main, final]{neurips_2026}

\usepackage[utf8]{inputenc} 
\usepackage[T1]{fontenc}    
\usepackage{hyperref}       
\usepackage{url}            
\usepackage{booktabs}       
\usepackage{amsfonts}       
\usepackage{nicefrac}       
\usepackage{microtype}      
\usepackage[table]{xcolor}  

\usepackage{amsmath} 
\usepackage{graphicx}
\usepackage{multirow}
\usepackage[capitalize,noabbrev]{cleveref}   
\crefname{lstlisting}{listing}{listings}
\Crefname{lstlisting}{Listing}{Listings}
\usepackage{caption}                          
\usepackage{xspace}
\graphicspath{{./}}
\usepackage{abbrv}

\usepackage{subcaption}
\usepackage{enumitem}
\usepackage{listings}

\usepackage{tikz}
\usetikzlibrary{positioning,fit,calc,shapes,arrows.meta,shadows.blur,backgrounds}
\definecolor{DarkBlue}{rgb}{0.05, 0.15, 0.35}
\definecolor{InvisibleBlue}{rgb}{0.92, 0.92, 0.97}
\definecolor{InvisibleGreen}{rgb}{0.92, 0.97, 0.92}
\definecolor{AccentGreen}{RGB}{39,174,96}
\definecolor{AccentOrange}{RGB}{230,126,34}
\definecolor{HighlightYellow}{RGB}{255,245,157}

\definecolor{promptbg}{RGB}{245,245,245}
\definecolor{promptframe}{RGB}{130,130,130}

\title{Driving Video Retrieval for Complex Queries with Structured Grounding}

\author{%
  \textbf{Manyi Yao}$^\dagger$, \textbf{Sparsh Garg}$^\ddagger$, \textbf{Christian Shelton}$^\dagger$,\\ \textbf{Amit Roy-Chowdhury}$^\dagger$, \textbf{Abhishek Aich}$^\ddagger$\thanks{Corresponding author: \href{mailto:aaich001@ucr.edu}{\texttt{aaich001@ucr.edu}}.}\\
$^\ddagger$NEC Laboratories, America, $^\dagger$University of California, Riverside
}

\begin{document}

\maketitle

\begin{abstract}
Video retrieval at scale is central to data curation and safety validation in autonomous driving, where users want to find not only scenes but also dynamic events such as cut-ins and hard braking. Existing vision-language and keyword-based retrieval methods often miss these events because the relevant motion may not be explicitly described in text or captured by lexical overlap. Rule-based retrieval can encode such events more directly, but it is brittle: generated or hand-written rules often fail when their assumptions do not match real driving data.
We propose {\ours}, a data-calibrated retrieval framework for driving videos. It uses weakly labeled in-domain videos to estimate when a query rule is reliable, adapt rules that mismatch observed data, and fuse calibrated rule scores with vision-language and keyword-based retrieval signals. Across three driving benchmarks, including newly released human-annotated event data on DrivingDojo, {\ours} delivers up to $84\%$ relative improvement in top-1 accuracy over state-of-the-art method, and remains effective under simulated degraded perception. Annotations are available at \url{https://github.com/nec-labs-ma/strive-d-data-annotation}.

\end{abstract}

\section{Introduction}
\label{sec:intro}

The deployment of advanced driver-assistance and autonomous systems has put fleet-scale dashcam and sensor logs at the center of safety engineering, with a single instrumented vehicle generating tens of terabytes of footage per day~\cite{yu2020bdd100k,caesar2020nuscenes,sun2020waymo}. Within these archives, the events that matter most for safety, such as rare interactions, near-misses, and abnormal driving behaviors, are sparsely distributed and difficult to surface manually. Natural-language video retrieval is the principal tool engineers use to query these archives by intent.

\begin{figure}[tb]
  \centering
  \includegraphics[width=0.99\textwidth, trim=0 135 30 0, clip]{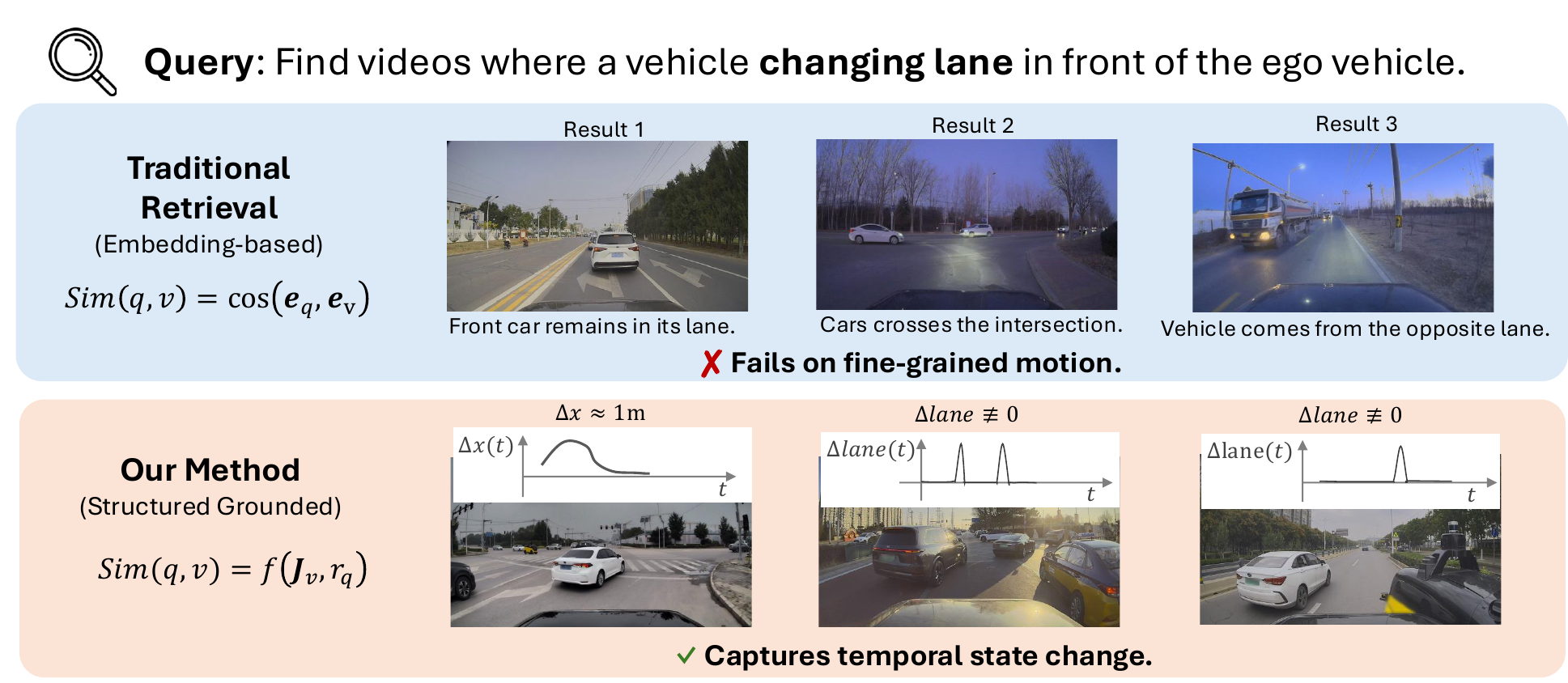}
  \caption{\textbf{Advantages of our method.} Embedding-based retrieval captures global scene similarity but often ignores short-term motion dynamics, such as lane changing and hard braking. Our structured grounding method explicitly models these motion patterns, enabling accurate retrieval of such events.}
  \label{fig:teaser}
\end{figure}

The dominant approach uses dense vision--language embeddings~\cite{tschannen2025siglip,qwen3vlembedding} to score query--video similarity. These embeddings pool fine-grained kinematic patterns, such as a hard brake or a lane cut-in, into a single representation that mixes them with surrounding scene content; queries that hinge on specific motion patterns are often not reliably answered by embedding similarity alone. We refer to this failure mode as \textbf{dilution} (\Cref{fig:teaser}).

Structured-grounding approaches~\cite{choi2024towards,shah2025neus,suris2023vipergptvisualinferencepython,liang2024neuralsymbolicvideoqalearningcompositional} address dilution by translating queries into executable rules over per-frame perception output and scoring videos against the rule rather than against a single embedding. Rule-based scoring depends on a rule's numerical constraints: thresholds at which gates fire, magnitudes that normalize graded scores, window lengths, and combination weights. A large language model (LLM) at synthesis time has text-level priors over what events ``mean,'' but those priors do not reflect the empirical distribution of signals as measured by the deployed perception pipeline; the numerical constraints an LLM proposes are therefore typically structurally plausible but miscalibrated in scale. We refer to this as \textbf{miscalibration}. Prior structured-retrieval methods either hand-author rules with hand-set constraints or translate queries via LLM without empirically calibrating them against the operating distribution; closing this calibration gap is our central technical contribution. We reframe constraint calibration as learning-to-rank under distant supervision~\cite{liu2009learning,mintz2009distant} (using weak supervision from auto-generated captions of unlabeled auxiliary videos), with an LLM proposing constraint assignments inside an iterative optimization loop~\cite{yang2024opro,romera2024funsearch}.

We introduce {\ours} (\textbf{STR}uctured and semant\textbf{I}c \textbf{V}ideo r\textbf{E}trieval, \textbf{D}riving), that instantiates structured-grounding retrieval around three components. First, a calibrated rule library handles queries that match library events directly; each library entry is an (event, rule) pair whose numerical constraints are fitted on auxiliary footage via the calibration loop above. Second, because the library covers a finite set of events while user queries may name events outside that inventory (a \textbf{coverage} gap), a confidence-aware adaptation policy decides per query whether to reuse a calibrated rule or synthesize a new rule conditioned on the closest calibrated rule as an in-context exemplar. Third, the symbolic score is combined with dense and lexical retrieval through query-conditioned routing, complementing the symbolic path on descriptive content and rare entities.

The primary contributions of this work are summarized as follows:
\begin{itemize}
  \item \textbf{Failure-mode reframe.} We characterize three distinct failure modes of structured-grounding retrieval over driving footage: \textbf{dilution} of motion signals in dense embeddings, \textbf{miscalibration} of LLM-proposed numerical constraints against the deployed perception pipeline, and \textbf{coverage} gaps in any fixed rule library. We design a method whose three components each address one.
  \item \textbf{Numerical-constraint calibration via learning-to-rank under distant supervision.} An LLM-as-optimizer loop jointly fits a rule's thresholds, magnitudes, and window lengths against auxiliary captions; no labels are required in the target benchmark.
  \item \textbf{Confidence-aware adaptation policy.} For queries outside the calibrated library, the policy bridges calibrated reuse and exemplar-conditioned synthesis, preserving the calibrated parameter scale while allowing structural specialization.
  \item \textbf{Empirical receipts and benchmark release.} On three driving benchmarks, {\ours} matches or exceeds the strongest dense and structured baselines on every metric (\eg, $83.6\%$ relative Acc@1 gain on DrivingDojo~\cite{wang2024drivingdojodatasetadvancinginteractive}, $\sim  1{,}570\times$ lower per-query latency than NSVS-TL~\cite{choi2024towards}); we release a human-annotated event-grounding set on DrivingDojo.
\end{itemize}

\section{Related Works}
\label{sec:works}
\paragraph{Vision-Language and Lexical Video Retrieval.}
Dense vision--language models~\cite{qwen3vlembedding,tschannen2025siglip,wang2025internvl3} retrieve videos by similarity in a shared video--text embedding space, with retrieval-focused variants adding temporally aware aggregation~\cite{xu2026hitea,luo2021clip4clipempiricalstudyclip,ma2022xclipendtoendmultigrainedcontrastive}. Recent video LLMs and retrieval-augmented systems further integrate generation into the same representation~\cite{damonlpsg2024videollama2,li2023videochat,jeong2025videoragretrievalaugmentedgenerationvideo}. Sparse lexical retrieval~\cite{robertson2009probabilistic,formal2021spladesparselexicalexpansion} complements these by matching query terms against indexed captions on rare-entity and long-tail queries. Recent benchmarks~\cite{kim2025videocompadvancingfinegrainedcompositional,Ventura2024TFCoVR,shen2025reasoningtexttovideoretrievaldigital} document a persistent shortfall of both paradigms on queries that require fine-grained spatiotemporal compositionality. Our work addresses this shortfall for motion-event retrieval, complementing dense and lexical retrieval with a structured rule-based path (\Cref{sec:method}).

\paragraph{Neuro-Symbolic Video Understanding.}
Neuro-symbolic approaches to video understanding ground language queries in structured representations of the scene, executing rules or programs to verify physical conditions. Temporal-logic methods reason about long-form events by composing per-frame vision-language model (VLM) scores under a temporal logic specification~\cite{choi2024towards,shah2025neus}; and scene-graph reasoning composes spatial and temporal predicates over parsed entities~\cite{liang2024neuralsymbolicvideoqalearningcompositional}. In the driving domain, related work translates dashcam content into structured representations for downstream language reasoning and analytics~\cite{Yao2025iFinder,pmlr-v288-leung25a,yan2025avaagenticvideoanalytics,sima2025drivelmdrivinggraphvisual,wen2024diluknowledgedrivenapproachautonomous}. Across these methods, the numerical constraints that the underlying LLM emits (e.g., thresholds on distance changes or deceleration) are typically used as-is, without an explicit calibration step against the operating distribution. Our work introduces such a calibration step, leaving the symbolic translation to the LLM and grounding the numerical constraints in auxiliary in-domain data.

\paragraph{Weak Supervision, Ranking, and LLM-as-Optimizer.}
Our calibration procedure draws on three established lines of work. Learning to rank under binary relevance~\cite{liu2009learning} provides the standard formulation for optimizing average precision over positive/negative pairs. Distant supervision~\cite{mintz2009distant} obtains training pairs from heuristic alignments rather than human labels, a paradigm widely adopted in natural-language processing and information retrieval. Recent work casts large language models as discrete optimizers, using their proposal distribution to search structured solution spaces under black-box objective evaluations~\cite{yang2024opro,romera2024funsearch}. Our method combines these three traditions in the setting of symbolic video retrieval: positive/negative video pairs derived from auxiliary captions provide weakly supervised rankings, and an LLM-driven proposal loop searches over rule parameters under the average-precision objective.

%

\section{Method}
\label{sec:method}

\subsection{Problem Formulation}
\label{sec:problem}
We address natural-language video retrieval. Given a query $\query$ and a corpus of $\numvideos$ videos $\corpus = \{\video_1, \ldots, \video_\numvideos\}$, the task is to learn a ranking function $\rankfn$ that returns the top-$\topK$ most relevant videos. We do \emph{not} assume access to query--video relevance labels on the target corpus. The only labels available during training are weak signals derived from a separate auxiliary set of in-domain footage; their construction is described in \Cref{sec:library}.

\begin{figure}[tb]
  \centering
  \includegraphics[width=0.99\textwidth]{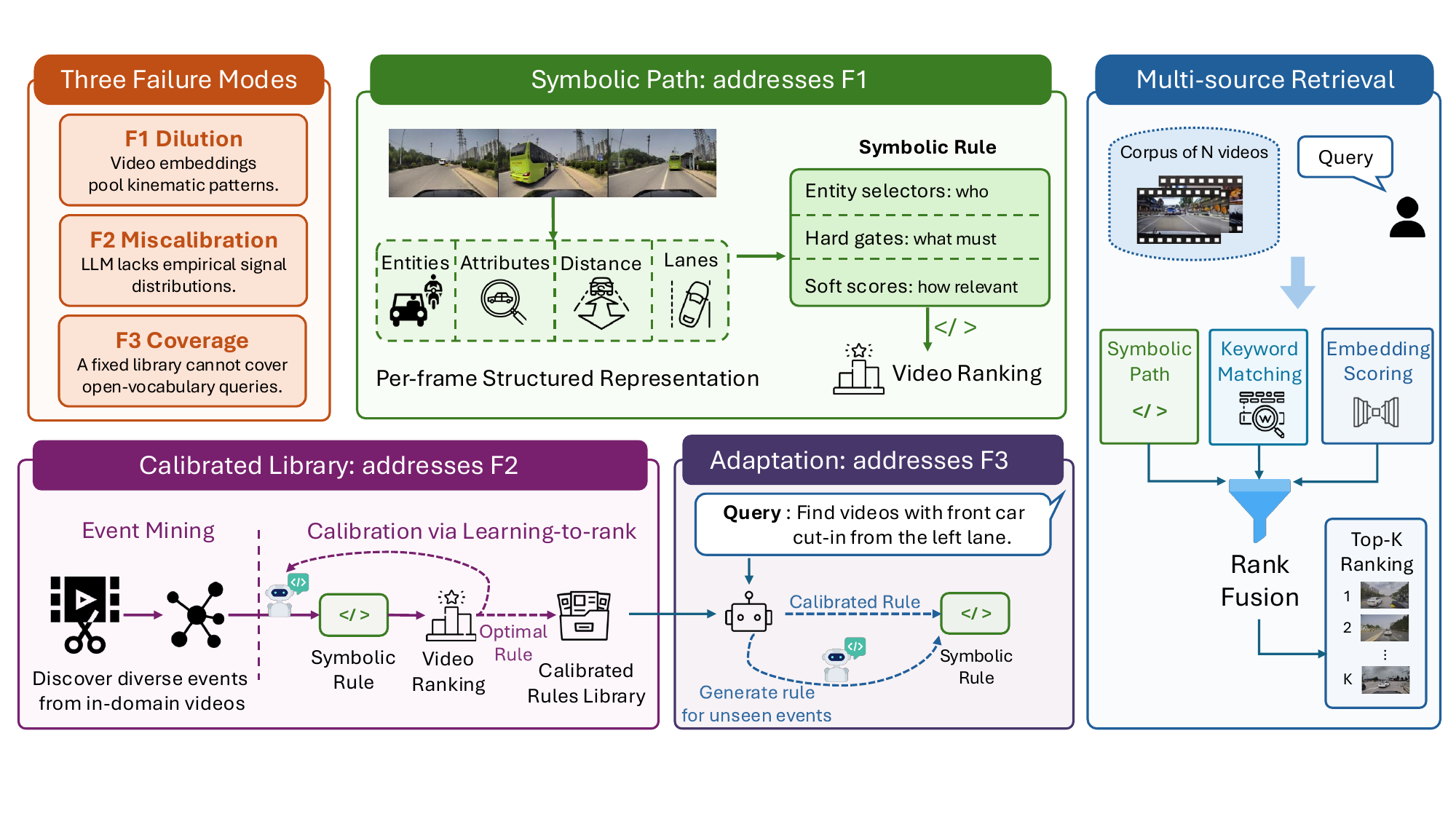}
  \caption{\textbf{Overview of \ours.} We identify three failure modes of existing query-to-video retrieval in driving (\textit{top-left}): F1 \textit{Dilution} of kinematic patterns by dense embeddings, F2 \textit{Miscalibration} of LLM-generated rules against empirical signal distributions, and F3 limited \textit{Coverage} of any fixed rule library for queries beyond the library's event inventory. Three components address them in turn: the \textit{Symbolic Path} (\textit{top-middle}, F1) scores videos by matching a per-frame structured representation (entities, attributes, distance, lane) against a symbolic rule composed of entity selectors, hard gates, and soft scores; the \textit{Calibrated Library} (\textit{bottom-left}, F2) mines diverse events from in-domain videos and calibrates rule thresholds via learning-to-rank, yielding a library of empirically grounded rules; \textit{Adaptation} (\textit{bottom-middle}, F3) handles unseen queries at test time by generating new rules from the calibrated library. \textit{Multi-source Retrieval} (\textit{right}) combines the symbolic path with keyword matching and embedding scoring through rank fusion to produce the final top-$K$ ranking.}

  \label{fig:main_framework}
\end{figure}

\subsection{Three Failure Modes}
\label{sec:overview}
Motion-event retrieval over driving footage exhibits three failure modes that, in our experiments, no single retrieval mechanism resolves on its own. \textbf{(F1) Dilution.} Dense vision--language embeddings pool fine-grained kinematic patterns (e.g., a hard brake, a lane cut-in) into a single representation that mixes them with surrounding scene content; queries that hinge on specific motion patterns are therefore often not reliably answered by embedding similarity alone. \textbf{(F2) Miscalibration.} Symbolic rules over per-frame perception output mitigate dilution by probing physical signals directly, but their \emph{numerical thresholds} (the values at which a rule fires) depend on the empirical distribution of those signals, to which an LLM at synthesis time has no access. The thresholds an LLM produces therefore tend to reflect text-level priors rather than the operating distribution, leaving rules miscalibrated in scale. \textbf{(F3) Coverage.} Even a fully calibrated library of rules is necessarily incomplete: query-time intents are not restricted to the library's event inventory, and any fixed library will miss matches.

{\ours} answers these failure modes with three components, illustrated in \Cref{fig:main_framework}. A \emph{symbolic path} (\Cref{sec:symbolic}) translates queries into executable symbolic rules over per-frame structured representation and ranks videos by satisfaction, addressing \textbf{F1}. A \emph{calibrated library} (\Cref{sec:library}), constructed once on auxiliary in-domain footage, fits the numerical constraints of each rule via learning-to-rank (L2R) under distant supervision, addressing \textbf{F2}. A \emph{confidence-aware adaptation policy} operates within the symbolic path: at retrieval time, it reuses a calibrated rule when the query semantically matches a library event, and otherwise synthesizes a new rule conditioned on the closest calibrated rule as an in-context exemplar, switching between calibrated reuse and exemplar-conditioned synthesis to address \textbf{F3}. Finally, \emph{multi-source fusion} (\Cref{sec:fusion}) combines the symbolic path with dense and sparse retrieval signals (defined in \Cref{sec:fusion}), complementing the symbolic rules on descriptive and rare-entity content that they are not designed to express.

\subsection{Symbolic Path: Rule Design}
\label{sec:symbolic}

\paragraph{Per-frame structured representation.}
Each video $\video$ in the target corpus is represented as a per-frame structured record $\vjson\video$ produced by an upstream perception pipeline operating on raw video. Each frame contains a set of tracked objects with associated fields such as object class, distance from the ego vehicle, lane assignment, and lateral offset. We use the structured representation as input to scoring. The specific perception components and field schema used in our experiments are described in \Cref{sec:exps} and \Cref{sec:settings}.

\paragraph{Rule composition.}
A motion-event description naturally decomposes into three questions: \emph{who} is involved, \emph{what} qualitative conditions must hold, and \emph{how strongly} the relevant fields change within a short window. A symbolic rule captures these as three components: \textbf{entity selectors}, \textbf{hard gates}, and \textbf{soft scores}, respectively. A rule's overall score combines them: hard gates determine admissibility (a video scores zero if any single precondition is unsatisfied), soft scores provide the graded ranking magnitude (a calibrated weighted average across multiple constraints), and the entity selector aggregates per-object scores to a video-level score (\textit{any}-mode for events localized to a single agent, \textit{consensus}-mode for group behaviors). Two design properties make the decomposition useful in practice. The \emph{Boolean-plus-continuous} split between hard gates and soft scores mirrors the satisfaction-and-robustness semantics of signal temporal logic~\cite{maler2004stl,donze2010robust}, and yields an interpretable composite (a video's relevance is its temporal-pattern strength conditioned on the event's preconditions being met) rather than a single opaque scalar. The split also keeps the per-event constraint set small and flat: a handful of thresholds, magnitudes, window lengths, and weights, so the calibration loop in \Cref{sec:library} can search the constraint space jointly per event. \Cref{lst:dsl-cutin} shows a calibrated rule for the event \emph{cut-in}; a per-component ablation isolating the contribution of each piece is in \Cref{sec:dd_composition}.

\begin{lstlisting}[caption={A calibrated symbolic rule for the event \emph{cut-in}. The schema and operators are fixed; the numerical values \texttt{40.0}, \texttt{4.0}, and \texttt{14.0} are fit by L2R against an auxiliary set (\Cref{sec:library}).}, label={lst:dsl-cutin}]
{
  "entity":      { "class_filter": ["car","truck"], "agg_mode": "any" },     // who
  "hard_gates":  [ { "field": "distance", "op": "<", "value": 40.0 } ],     // what must
  "soft_scores": [ { "field": "loc_x", "type": "max_change",
                     "window_seconds": 4.0, "magnitude": 14.0 } ]           // how strongly
}
\end{lstlisting}

\subsection{Calibrated Library: Construction and Adaptation}
\label{sec:library}

\paragraph{Event mining.}
For each video $\bvideo$ in a separate auxiliary set of in-domain footage (disjoint from the target corpus and unlabeled), a video captioner generates a description $\bvcap$ targeted at kinematic content rather than generic scene description. An LLM then extracts candidate motion events from each $\bvcap$ as (event type, object) pairs, and consolidates them across videos into a deduplicated set $\sevent = \{\bevent_1, \ldots, \bevent_\librarysize\}$. For each event $\beventi$, an auxiliary video is treated as positive if the LLM extracts $\beventi$ from its caption, and as negative otherwise; the resulting sets $\vpos, \vneg$ serve as weak supervision. We rely on captions only to label which auxiliary videos belong to $\vpos$ versus $\vneg$; candidate rules are scored against the auxiliary videos' per-frame structured representation, not their captions.

\paragraph{Calibration via learning-to-rank.}
For each event $\beventi$, an LLM constructs a rule within a shared, manually defined schema, conditioned on the event label and auxiliary-set field statistics. We hold the resulting structure fixed while optimizing its numerical constraints~\cite{yang2024opro,romera2024funsearch}. The loop alternates between three steps: (i) the LLM proposes a candidate constraint assignment for the rule; (ii) the assignment is scored by the average precision (AP) of the ranking it induces over $\vpos \cup \vneg$, computed from the auxiliary videos' per-frame structured representation; (iii) the LLM is shown the AP it just achieved together with the best-scoring assignment from prior iterations and asked to propose a revision. The loop terminates when AP exceeds a target threshold or after $\nrev$ iterations, and the best assignment populates the calibrated library $\library$. AP under distant supervision~\cite{liu2009learning,mintz2009distant} is the sole selection criterion: captions are a noisy motion-event signal (captioners often emphasize scene description over kinematic detail), but AP averages over many positive and negative candidates per event, mitigating this noise. Direct caption-based retrieval offers no comparable averaging, since each query depends on a single per-video score, so motion events not verbalized in the caption tend to be missed.

\paragraph{Adaptation at retrieval time.}
The calibrated library is, by construction, incomplete: queries may not match any library event. We use a confidence-aware policy that bridges between calibrated reuse and on-the-fly synthesis. For a query $\query$, an LLM identifies the closest event $\bestevent \in \sevent$ and produces a confidence score $\mconf \in [0, 1]$. When $\mconf \ge \tauh$, the calibrated rule is reused. Otherwise, the LLM adapts the closest calibrated rule to the new query, treating it as an in-context exemplar; this preserves the calibrated threshold scale of the related event while allowing structural specialization.

\subsection{Multi-Source Fusion}
\label{sec:fusion}
The symbolic path is designed for physically grounded motion events. Descriptive content and rare entities are better served by two complementary retrieval signals: \emph{dense retrieval}, which scores videos by similarity in a shared video--text embedding space, and \emph{sparse retrieval}, which matches query terms against per-video captions via lexical overlap. The three sources are complementary rather than substitutes for each other. To merge their per-source rankings into a single ordering, we use a query-conditioned weighted reciprocal rank fusion~\cite{10.1145/1571941.1572114} in which the per-source weights are produced by an LLM that we call the \emph{fusion router}: given the query and a short description of each source's strengths, the fusion router outputs a weight per source, and these weights scale the rank-based contribution of each source in the fusion.

%
\section{Experiments}
\label{sec:exps}

\subsection{Settings, Datasets, and Metrics}
\label{sec:exp_settings}

\paragraph{Perception pipeline and library construction.}
Per-frame structured representations (\Cref{sec:symbolic}) are produced by iFinder's perception pipeline~\cite{Yao2025iFinder}, with fields including class, distance-to-ego, lane assignment, and lateral offset. All perception modules use their released configurations without dataset-specific tuning. The auxiliary set is the Nexar corpus~\cite{moura2025nexardashcamcollisionprediction}; from $1{,}147$ Nexar videos disjoint from any evaluation benchmark, we mine and calibrate a rule library of $\librarysize=21$ events. No queries, labels, or videos from evaluation benchmarks enter library construction.

\paragraph{Models and hyperparameters.}
All language model components (rule synthesis, the library matcher used by the confidence-aware adaptation policy, and multi-source fusion weighting) use GPT-4.1~\cite{openai2025gpt41} unless otherwise specified, with full prompts in \Cref{sec:prompts}. Dense retrieval uses Qwen3-VL-Embedding-8B~\cite{qwen3vlembedding}; sparse retrieval uses BM25~\cite{robertson2009probabilistic} over auxiliary captions. The reranking pool size is fixed at $\npool=20$ for all methods that include a reranker stage; this value is selected from the per-method pool sweep in \Cref{sec:pool_size}, where it is at or near the optimum for every evaluated model. The library matcher confidence threshold is $\tauh=0.6$, the L2R proposal loop runs up to $\nrev=20$ iterations per event, and the RRF smoothing constant $k_{\mathrm{RRF}}=60$ (following~\cite{10.1145/1571941.1572114}). All experiments run on a single NVIDIA RTX A6000 (48~GB).

\paragraph{Datasets and queries.}
We evaluate on three dashcam corpora that exercise complementary query types.
\textit{DrivingDojo}~\cite{wang2024drivingdojodatasetadvancinginteractive} contains 583 videos; we release human annotations of 41 distinct motion events tied to specific object classes (\eg, ``\textit{truck cut in},'' ``\textit{ego car change lane}''), supporting fine-grained event-grounded retrieval; 236 videos are relevant to at least one query (\Cref{sec:drivingdojo-annotations}). 
\textit{CarCrashDataset}~\cite{BaoMM2020} provides ``reason of accident'' annotations across 45 categories (\eg, ``\textit{slipping in the snow}''); we evaluate on the 110 videos carrying these annotations, targeting causal/explanatory queries.
\textit{MM-AU}~\cite{Fang_2024_CVPR} contains $1{,}953$ test videos with 49 accident-type annotations (\eg, ``\textit{ego car hits a crossing pedestrian}''), targeting accident-description queries. The three datasets together span queries from compositional motion (DrivingDojo), through causal explanation (CarCrash), to accident description (MM-AU).

\paragraph{Baselines.}
We compare against three families: (i) \emph{embedding-based} dual encoders, including Qwen3-VL-Embedding (2B/8B)~\cite{qwen3vlembedding}, SigLIP~\cite{zhai2023sigmoidlosslanguageimage}, and SigLIP2~\cite{tschannen2025siglip} (Qwen3-VL is reported paired with its reranker; without-reranker variants in \Cref{sec:additional_exps}); (ii) \emph{retrieval-augmented} VideoRAG~\cite{jeong2025videoragretrievalaugmentedgenerationvideo}; and (iii) \emph{neuro-symbolic} NSVS-TL~\cite{choi2024towards}. All baselines are used in their released form;  \Cref{sec:structured_grounding} substitutes NSVS-TL into our pipeline with identical dense, sparse, router, and reranker components.

\paragraph{Evaluation metrics.}
We report top-$k$ accuracy ($k\in\{1,3,5,10\}$), the fraction of queries with at least one ground-truth video in the top-$k$; mean reciprocal rank (MRR); and mean average precision (mAP). We do not report recall because the number of relevant videos per query varies widely across our annotations, making recall not directly comparable across queries.

\subsection{Main Results}
\label{sec:main_results}

\begin{table}[!t]
 \centering
 \small
 \setlength{\tabcolsep}{4pt}
 \caption{\textbf{Quantitative comparison on three benchmarks (\%).} All methods are evaluated under a unified reranking pool size $\npool=20$. {\ours} achieves the highest score on every metric on every dataset. \textbf{Bold}: best per column. \underline{Underline}: second-best.}
 \label{tab:main_results}
 \begin{tabular}{clcccccc}
 \toprule
 Type & Method (Year) & MRR($\uparrow$) & mAP($\uparrow$) & Acc@1 & Acc@3 & Acc@5 & Acc@10 \\
 \specialrule{\lightrulewidth}{\aboverulesep}{0pt}
 \rowcolor{gray!15}
 \multicolumn{8}{c}{\textbf{\textit{DrivingDojo}}\cite{wang2024drivingdojodatasetadvancinginteractive}} \\
 \midrule
 \multirow{4}{*}{\textit{Embedding}} & QWen3-vl-2B\cite{qwen3vlembedding}(2026) & 12.2 & 1.4 & 4.9 & 26.8 & 46.3 & 65.9 \\
 & QWen3-vl-8B\cite{qwen3vlembedding}(2026) & 20.1 & 2.2 & 14.6 & 31.7 & \underline{51.2} & \underline{68.3} \\
 & SigLip\cite{zhai2023sigmoidlosslanguageimage}(2023) & 23.7 & 4.7 & 14.6 & 22.0 & 31.7 & 58.5 \\
 & SigLip2\cite{tschannen2025siglip}(2025) & 25.0 & 1.9 & 14.6 & 31.7 & 34.2 & 48.8 \\
 \cmidrule(lr){1-8}
 \textit{RAG} & VideoRAG\cite{jeong2025videoragretrievalaugmentedgenerationvideo}(2025) & 21.6 & 3.5 & 12.2 & 29.3 & 29.3 & 39.0 \\
 \cmidrule(lr){1-8}
 \multirow{2}{*}{\textit{Neuro-sym.}} & NSVS-TL\cite{choi2024towards}(2024) & \underline{31.8} & \underline{4.9} & \underline{22.0} & \underline{34.2} & 43.9 & 53.7 \\
 & {\ours} (Ours) & \textbf{38.7} & \textbf{8.6} & \textbf{26.8} & \textbf{48.8} & \textbf{53.7} & \textbf{73.2} \\
 \specialrule{\heavyrulewidth}{4pt}{0pt}
 \rowcolor{gray!15}
 \multicolumn{8}{c}{\textbf{\textit{MM-AU}}~\cite{Fang_2024_CVPR}} \\
 \midrule
 \multirow{4}{*}{\textit{Embedding}} & QWen3-vl-2B\cite{qwen3vlembedding}(2026) & 26.9 & 6.0 & 18.4 & 30.6 & 42.9 & 49.0 \\
 & QWen3-vl-8B\cite{qwen3vlembedding}(2026) & \underline{32.7} & \underline{6.3} & \underline{20.4} & \underline{38.8} & \underline{55.1} & \underline{61.2} \\
 & SigLip\cite{zhai2023sigmoidlosslanguageimage}(2023) & 16.1 & 4.3 & 6.1 & 24.5 & 30.6 & 38.8 \\
 & SigLip2\cite{tschannen2025siglip}(2025) & 17.8 & 6.1 & 8.2 & 22.5 & 26.5 & 40.8 \\
 \cmidrule(lr){1-8}
 \textit{RAG} & VideoRAG\cite{jeong2025videoragretrievalaugmentedgenerationvideo}(2025) & 22.9 & 6.0 & 18.4 & 24.5 & 24.5 & 36.7 \\
 \cmidrule(lr){1-8}
 \multirow{2}{*}{\textit{Neuro-sym.}} & NSVS-TL\cite{choi2024towards}(2024) & 18.0 & 2.8 & 6.1 & 20.4 & 36.7 & 53.1 \\
 & {\ours} (Ours) & \textbf{40.5} & \textbf{9.5} & \textbf{26.5} & \textbf{53.1} & \textbf{59.2} & \textbf{65.3} \\
 \specialrule{\heavyrulewidth}{4pt}{0pt}
 \rowcolor{gray!15}
 \multicolumn{8}{c}{\textbf{\textit{CarCrashDataset}}~\cite{BaoMM2020}} \\
 \midrule
 \multirow{4}{*}{\textit{Embedding}} & QWen3-vl-2B\cite{qwen3vlembedding}(2026) & 38.3 & 32.1 & \underline{26.7} & 44.4 & 55.6 & 64.4 \\
 & QWen3-vl-8B\cite{qwen3vlembedding}(2026) & \underline{40.1} & \underline{33.5} & \textbf{28.9} & \underline{46.7} & \underline{62.2} & \underline{66.7} \\
 & SigLip\cite{zhai2023sigmoidlosslanguageimage}(2023) & 30.0 & 22.0 & 17.8 & 35.6 & 51.1 & 60.0 \\
 & SigLip2\cite{tschannen2025siglip}(2025) & 28.9 & 23.4 & 15.6 & 33.3 & 51.1 & 62.2 \\
 \cmidrule(lr){1-8}
 \textit{RAG} & VideoRAG\cite{jeong2025videoragretrievalaugmentedgenerationvideo}(2025) & 32.1 & 25.6 & 20.0 & 40.0 & 46.7 & 55.6 \\
 \cmidrule(lr){1-8}
 \multirow{2}{*}{\textit{Neuro-sym.}} & NSVS-TL\cite{choi2024towards}(2024) & 23.1 & 19.3 & 11.1 & 24.4 & 37.8 & 53.3 \\
 & {\ours} (Ours) & \textbf{45.0} & \textbf{37.7} & \textbf{28.9} & \textbf{60.0} & \textbf{73.3} & \textbf{75.6} \\
 \bottomrule
 \end{tabular}
\end{table}

\Cref{tab:main_results} reports retrieval performance on all three benchmarks under a unified reranking pool size $\npool=20$. {\ours} is highest on every metric on every dataset; the Acc@1 advantage over the neuro-symbolic baseline NSVS-TL~\cite{choi2024towards} ranges from $+4.8$ percentage points (pp) on DrivingDojo~\cite{wang2024drivingdojodatasetadvancinginteractive} to $+20.4$\,pp on MM-AU~\cite{Fang_2024_CVPR}.

Gains concentrate at top-of-list precision. On DrivingDojo~\cite{wang2024drivingdojodatasetadvancinginteractive}, {\ours} delivers $26.8\%$ Acc@1 versus $14.6\%$ for the strongest dense baseline (an $83.6\%$ relative improvement), and $8.6\%$ mAP versus $4.7\%$ ($+82.9\%$ relative). The pattern holds elsewhere: MM-AU~\cite{Fang_2024_CVPR} Acc@1 improves $20.4 \to 26.5$ ($+29.9\%$); on CarCrashDataset~\cite{BaoMM2020}, Acc@1 ties the best baseline at $28.9\%$ while every other metric dominates. Broad-recall metrics (Acc@10) close more because reranker-equipped baselines partially recover at wider candidate pools, a behavior dense alone cannot deliver at top-of-list precision. Across five full-pipeline runs, the Acc@1 advantage over the strongest dense baseline persists on DrivingDojo and MM-AU (~\Cref{sec:error-analysis}).

\subsection{Ablations and Analysis}
\label{sec:dd_analysis}

\paragraph{Component ablation.}
\label{sec:dd_components}
\begin{figure}[!t]
  \centering
  \begin{minipage}[t]{0.48\linewidth}
    \centering
    \includegraphics[width=\linewidth]{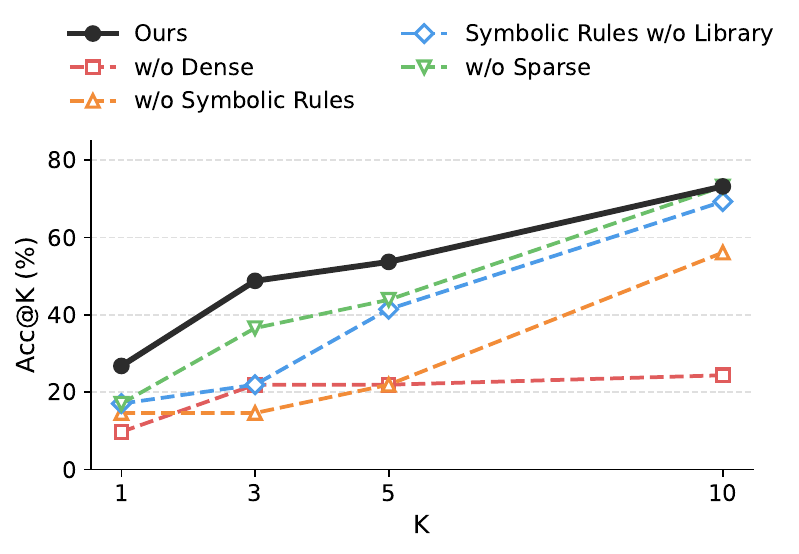}
    \captionof{figure}{\textbf{Component ablation on DrivingDojo.} Removing the dense path causes the most severe degradation; removing the calibrated library, symbolic-rule path, or sparse retrieval each isolates a distinct failure mode.}
    \label{fig:abllation_components}
  \end{minipage}
  \hfill
  \begin{minipage}[t]{0.48\linewidth}
    \centering
    \includegraphics[width=\linewidth]{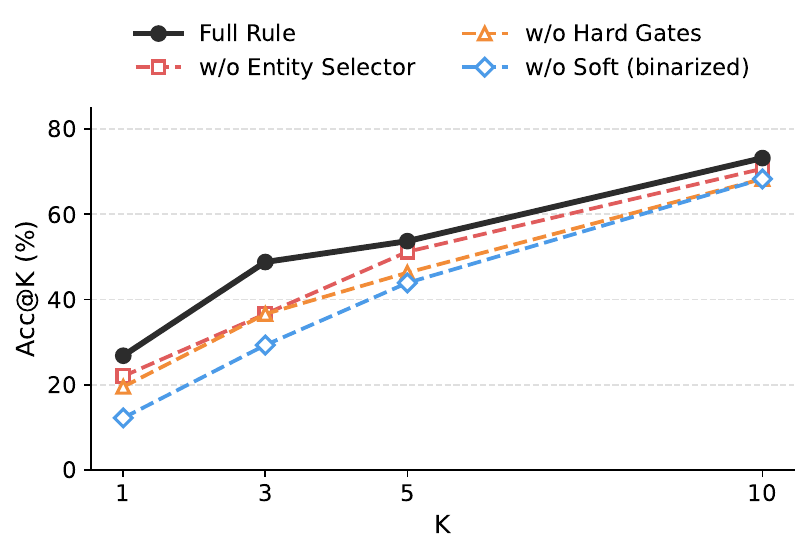}
    \captionof{figure}{\textbf{Rule composition ablation on DrivingDojo.} All three pieces are load-bearing. Soft scores carry the most weight at top-of-list precision; hard gates contribute substantially; entity selectors contribute incrementally.}
    \label{fig:dsl_composition}
  \end{minipage}
\end{figure}

\Cref{fig:abllation_components} removes each component of {\ours} in turn. \textbf{Removing the dense path} causes the largest drop, Top-1 from $26.8\%$ to $9.76\%$: sparse and symbolic rules alone do not fully recover the descriptive content carried by dense embeddings, and dense alone does not reach the same top-of-list precision (\Cref{tab:main_results}); the two pathways are complementary. \textbf{Removing the calibrated library} forces the LLM to synthesize rules from scratch at query time, using the same schema (entity selectors, hard gates, soft scores) constrained by few-shot exemplars but without L2R threshold calibration; Top-3 drops sharply with partial recovery at Top-10. Because the rule \emph{form} is held constant, this gap isolates threshold calibration from any structural variation. \textbf{Removing the symbolic path entirely} also drops Top-3 sharply, indicating that the symbolic path contributes precision that dense+sparse alone do not provide in this setting. \textbf{Removing sparse retrieval} leaves Top-10 essentially unchanged but slightly hurts top-rank precision, consistent with sparse's rare-entity recovery role rather than primary recall.

\paragraph{Rule composition: entity selectors, hard gates, soft scores.}
\label{sec:dd_composition}

Each symbolic rule (\Cref{sec:symbolic}) is composed of three pieces (entity selector, hard gates, soft scores) mapping to \emph{who}, \emph{what conditions}, and \emph{how strongly}. We disable each piece in turn while holding calibrated thresholds, fusion, and the reranker fixed (no L2R re-run).
\Cref{fig:dsl_composition} shows all three pieces are load-bearing, with a clear ordering at top-of-list precision. \textbf{Graded soft scores carry the most weight}: binarizing them at the calibrated threshold drops Acc@1 from $26.8\%$ to $12.2\%$ ($-14.6$\,pp), as the rule loses the ability to distinguish strong from marginal matches. \textbf{Hard gates contribute substantially} ($-7.3$\,pp Acc@1): the binary admissibility check filters out videos otherwise falsely admitted by soft scores alone. \textbf{The entity selector contributes incrementally} ($-4.8$\,pp Acc@1), the smallest drop, as dense and sparse fusion partially compensate for off-target object scoring. All variants converge at Acc@10 (within $5$\,pp), reflecting the dense+sparse recall floor; differentiation lives at top-of-list precision where the calibrated symbolic rule is most discriminative.

\paragraph{Plug-in generalizability.}
\label{sec:dd_plugin}
\begin{figure}[!t]
  \centering
  \includegraphics[width=0.99\textwidth]{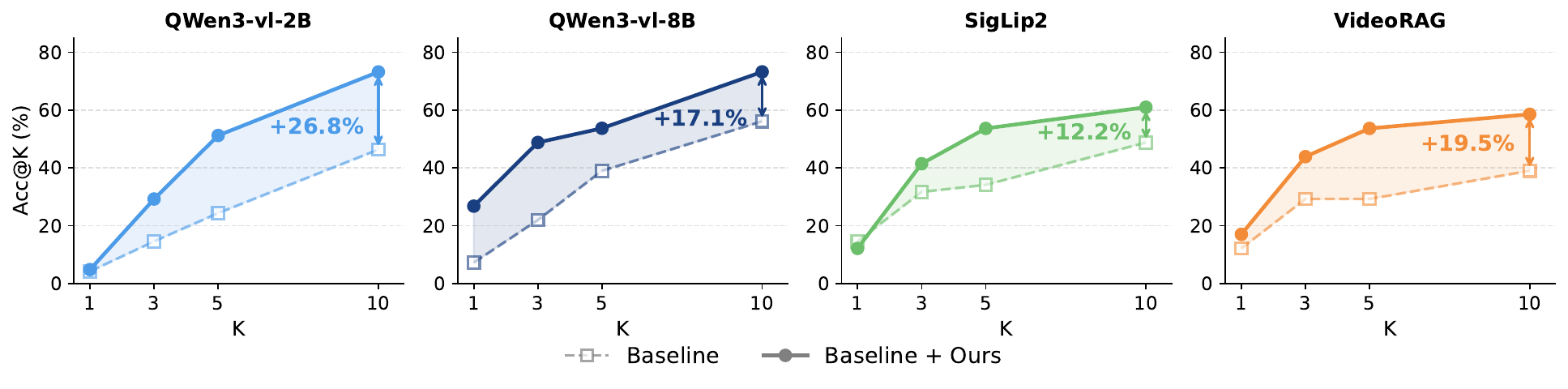}
  \caption{\textbf{Plug-in generalizability evaluated on DrivingDojo.} Our method consistently improves the full retrieval curve when integrated into 4 off-the-shelf retrievers spanning different architectures, from lightweight VLM embeddings (QWen3-vl\cite{qwen3vlembedding}) to video retrieval systems (VideoRAG\cite{jeong2025videoragretrievalaugmentedgenerationvideo}) and vision encoders (SigLip2\cite{tschannen2025siglip}). Numbers on the right indicate the absolute gain in top-10 accuracy.}

  \label{fig:plugin}
\end{figure}

We attach our symbolic-rule pathway to four off-the-shelf retrievers spanning architectures: Qwen3-VL-2B and -8B~\cite{qwen3vlembedding}, VideoRAG~\cite{jeong2025videoragretrievalaugmentedgenerationvideo}, and SigLIP2~\cite{tschannen2025siglip} (\Cref{fig:plugin}). Top-10 accuracy improves on every backbone, with gains ranging from $12.2$\,pp on SigLIP2 to $26.8$\,pp on Qwen3-VL-2B. The $17.1$\,pp lift on Qwen3-VL-8B is particularly informative: it shows that the improvement is not solely a result of compensation for a weak backbone.

\begin{table}[!t]
  \centering
  \begin{minipage}[t]{0.48\linewidth}
    \centering
\small
\setlength{\tabcolsep}{4pt}
\captionof{table}{\textbf{Online per-query latency on DrivingDojo (mean).} Embedding and retrieval-augmented methods run in milliseconds; structured-grounding methods run in seconds.}
\label{tab:latency}
\begin{tabular}{llr}
\toprule
Type & Method & Latency \\
\midrule
\multirow{2}{*}{\textit{Embedding}}
  & SigLIP2~\cite{tschannen2025siglip}            & $26.80$\,ms \\
  & Qwen3-VL-Emb-8B~\cite{qwen3vlembedding}       & $128.54$\,ms \\
\cmidrule(lr){1-3}
\textit{RAG}
  & VideoRAG~\cite{jeong2025videoragretrievalaugmentedgenerationvideo} & $37.18$\,ms \\
\midrule
\multirow{2}{*}{\textit{Neuro-sym.}}
  & {\ours} (Ours)                & $3.10$\,s \\
  & NSVS-TL~\cite{choi2024towards} & $4{,}868.05$\,s \\
\bottomrule
\end{tabular}

  \end{minipage}
  \hfill
  \begin{minipage}[t]{0.48\linewidth}
    \centering
\small
\setlength{\tabcolsep}{4pt}
\captionof{table}{\textbf{VLM binary classification baseline on DrivingDojo (\%).} Qwen3-VL-8B-Instruct~\cite{bai2025qwen3vl} is prompted with \textit{``Does this video show \{$\query$\}?''} per candidate; the +Reranker variant additionally re-ranks the predicted-\emph{Yes} subset with Qwen3-VL-Reranker-8B~\cite{qwen3vlembedding}.}
\label{tab:vlm_binary}
\begin{tabular}{lccc}
\toprule
Method & Acc@1 & Acc@5 & Acc@10 \\
\midrule
VLM Binary           & $2.4$ & $29.3$ & $41.5$ \\
~~+~Reranker         & $7.3$ & $34.2$ & $48.8$ \\
\midrule
{\ours}              & $\mathbf{26.8}$ & $\mathbf{53.7}$ & $\mathbf{73.2}$ \\
\bottomrule
\end{tabular}

  \end{minipage}
\end{table}

\paragraph{Efficiency.}
\label{sec:efficiency}
\Cref{tab:latency} reports mean per-query online latency on DrivingDojo~\cite{wang2024drivingdojodatasetadvancinginteractive}. {\ours} averages $3.10$\,s per query. Pure dense retrievers run in $\sim$$130$\,ms (Qwen3-VL-Embedding-8B~\cite{qwen3vlembedding}) but do not perform structured grounding and underperform on the precision metrics in \Cref{tab:main_results}. Within the structured-retrieval paradigm, the meaningful comparison is to NSVS-TL~\cite{choi2024towards}: at $4{,}868.05$\,s per query, NSVS-TL is $\sim$$1{,}500\times$ slower than {\ours}. Two LLM stages, rule synthesis (invoked only when no library rule matches the query) and multi-source fusion, account for $\sim$$87\%$ of the per-query cost, and both are amenable to caching across queries sharing an event category.

\paragraph{VLM binary classification baseline.}
\label{sec:vlm-binary}
A natural alternative to structured retrieval is to ask a general-purpose VLM directly: \textit{``Does this video show $\query$?''} for each candidate, and rank by the predicted \textit{Yes} probability. \Cref{tab:vlm_binary} reports this baseline on DrivingDojo using Qwen3-VL-8B-Instruct~\cite{bai2025qwen3vl}, with an optional Qwen3-VL-Reranker-8B~\cite{qwen3vlembedding} second stage. The scorer alone trails every baseline in \Cref{tab:main_results}, and even with the reranker it remains well below the strongest embedding-based baselines (Acc@1 $7.3$ vs.\ $14.6$). Per-video binary classification is not a competitive retrieval strategy on this benchmark.

\begin{table}[!t]
  \centering
  \begin{minipage}[t]{0.44\linewidth}
    \centering
\small
\setlength{\tabcolsep}{4pt}
\captionof{table}{\textbf{Calibration proposer comparison on DrivingDojo (Acc@10, \%)}. Calibrated variants share rule structure, AP objective, and budget (20 proposals per rule). $\pm$: std over 3 runs.}
\label{tab:proposer}
\begin{tabular}{lc}
\toprule
Proposer & Acc@10 \\
\midrule
Uncalibrated (no search) & $65.9$ \\
Random search            & $69.9 \pm 3.7$ \\
Bayesian search          & $69.9 \pm 2.8$ \\
Grid search              & $70.7$ \\
LLM (ours)               & $\mathbf{73.2}$ \\
\bottomrule
\end{tabular}

  \end{minipage}
  \hfill
  \begin{minipage}[t]{0.52\linewidth}
    \centering
\small
\setlength{\tabcolsep}{4pt}
\captionof{table}{\textbf{Library size vs.\ adaptation.} The calibrated library is reduced to random subsets of the 21 events (5 seeds per size). Adaptation rate: fraction of DrivingDojo queries routed to rule synthesis.}
\label{tab:library_size}
\begin{tabular}{ccc}
\toprule
\# library events & Adapt.\ rate (\%) & Acc@10 (\%) \\
\midrule
0 (no library) & $100$  & $69.3 \pm 0.0$ \\
5              & $85.4$ & $70.2 \pm 2.0$ \\
10             & $36.6$ & $71.7 \pm 1.3$ \\
21 (full)      & $12.2$ & $\mathbf{73.2 \pm 0.0}$ \\
\bottomrule
\end{tabular}

  \end{minipage}
\end{table}

\paragraph{Calibration proposer.}
\label{sec:proposer_ablation}
To assess the effect of the calibration proposer, we compare LLM and non-LLM search with fixed rule forms, auxiliary-set AP objective, and budget of 20 proposals per rule (\Cref{tab:proposer}). The LLM conditions proposals on per-field statistics and best-so-far feedback. All calibrated variants outperform the uncalibrated baseline ($65.9\%$ Acc@10). The LLM achieves $73.2\%$, compared with $70.7\%$ for the best non-LLM search, and also synthesizes query-specific rules when no library rule matches with sufficient confidence (\Cref{sec:library}).

\paragraph{Library size and query-time adaptation.}
\label{sec:library_size}
To assess the benefit of calibrated exemplars, we progressively reduce the 21-event library (\Cref{tab:library_size}). As the library shrinks from 21 events to zero, the fraction of queries routed to rule synthesis rises from $12.2\%$ to $100\%$, while Acc@10 decreases from $73.2\%$ to $69.3\%$. With all rules synthesized at query time, accuracy remains above the strongest dense baseline (Qwen3-VL-8B~\cite{qwen3vlembedding}, $68.3\%$). Calibrated exemplars improve accuracy, while query-time synthesis supports retrieval beyond the retained event inventory.

\paragraph{Robustness to perception errors.}
\label{sec:robustness}
To assess sensitivity to perception errors, we perturb precomputed records one module at a time, with the calibrated library and other retrieval paths fixed. Independent detection removal and track fragmentation reduce Acc@10 by at most $3.4$\,pp, remaining above the strongest dense baseline ($68.3\%$; \Cref{tab:robustness_summary}). Sustained missed detections and 20-frame track interruptions reduce accuracy to $62.6\%$ and $63.4\%$, respectively. Full protocols are in \Cref{sec:robustness_full}.

\begin{table}[t]
  \centering
  \begin{minipage}[t]{0.49\linewidth}
    \centering
\small
\setlength{\tabcolsep}{3pt}
\captionof{table}{\textbf{Perception robustness} (DrivingDojo; five-seed mean $\pm$ std). Rates are affected fractions; for false detections, added/original count ratios. Unperturbed Acc@10: $73.2\%$.}
\label{tab:robustness_summary}
\begin{tabular}{lc}
  \toprule
  Perturbation & Acc@10 (\%) \\
  \midrule
  \multicolumn{2}{l}{\emph{Independent (rate: $50\%$)}} \\
  Detection removal    & $70.2 \pm 2.0$ \\
  Track fragmentation  & $69.8 \pm 1.3$ \\
  Vehicle depth ($\times\mathcal{N}(1,0.5^2)$) & $72.2 \pm 3.3$ \\
  \midrule
  \multicolumn{2}{l}{\emph{Temporally correlated (rate: $75\%$)}} \\
  False detections               & $69.1 \pm 3.0$ \\
  Consecutive misses (per object) & $62.6 \pm 7.5$ \\
  Track interruption (20 frames)  & $63.4 \pm 3.1$ \\
  \bottomrule
\end{tabular}

  \end{minipage}
  \hfill
  \begin{minipage}[t]{0.47\linewidth}
    \centering
\small
\setlength{\tabcolsep}{3pt}
\captionof{table}{\textbf{Structured grounding choice} (DrivingDojo, \%).
NSVS-TL + pipeline replaces symbolic rules with NSVS-TL, with
dense retrieval, sparse retrieval, routing, and reranking fixed.}
\label{tab:nsvs_compare_main}
\begin{tabular}{lccc}
  \toprule
  Variant & Acc@1 & mAP & Acc@10 \\
  \midrule
  NSVS-TL standalone & $22.0$ & $4.9$ & $53.7$ \\
  NSVS-TL + pipeline & $12.2$ & $3.9$ & $70.7$ \\
  {\ours}            & $\mathbf{26.8}$ & $\mathbf{8.6}$ & $\mathbf{73.2}$ \\
  \bottomrule
\end{tabular}

  \end{minipage}
\end{table}

\paragraph{Structured grounding choice.}
\label{sec:structured_grounding}
To compare structured scoring methods, we substitute NSVS-TL for symbolic rules within the same retrieval pipeline. With the other components fixed, our rules improve Acc@1 by $14.6$\,pp and mAP by $4.7$\,pp (\Cref{tab:nsvs_compare_main}).

\subsection{Qualitative Results}
\label{sec:qual}
\begin{figure}[t]
  \centering
  \setlength{\tabcolsep}{2pt}
  \renewcommand{\arraystretch}{0.5}
  \newcommand{\labelours}{\rotatebox{90}{\small\textcolor{ForestGreen}{\textbf{Ours\hspace{6pt}}}}}
  \newcommand{\labelsiglip}{\rotatebox{90}{\small SigLIP2\hspace{4pt}}}
  \newcommand{\labelqwen}{\rotatebox{90}{\small Qwen3-8B\hspace{0pt}}}

  \begin{tabular}{c cc}
    & \small\textit{bus cut-in} & \small\textit{car collision with city guardrails} \\[3pt]
    \labelours &
    \includegraphics[width=0.44\textwidth]{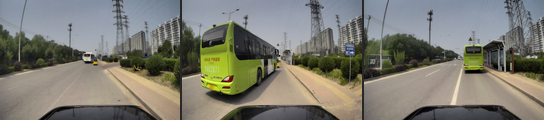} &
    \includegraphics[width=0.44\textwidth]{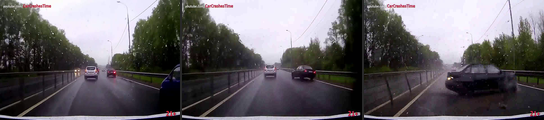} \\[3pt]
    \labelsiglip &
    \includegraphics[width=0.44\textwidth]{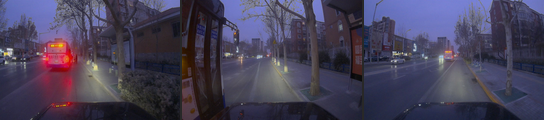} &
    \includegraphics[width=0.44\textwidth]{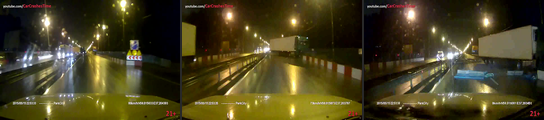} \\[3pt]
    \labelqwen &
    \includegraphics[width=0.44\textwidth]{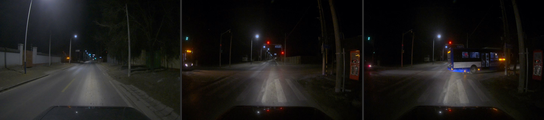} &
    \includegraphics[width=0.44\textwidth]{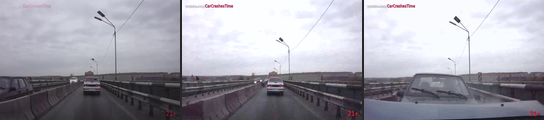} \\
  \end{tabular}

  \caption{\textbf{Qualitative retrieval comparison.} Each row shows the top retrieved results per model for two queries. For \textit{bus cut-in}, baselines retrieve videos containing buses but miss the cut-in motion. For \textit{car collision with city guardrails}, baselines retrieve either collisions without guardrails or guardrails without collisions. {\ours} correctly identifies both the object and the event in each case.}
  \label{fig:qualitative}
\end{figure}

\Cref{fig:qualitative} illustrates representative retrieval results. For the ``\textit{bus cut-in}'' query, baseline models retrieve videos containing buses but fail to identify the cut-in behavior, returning scenes where buses are simply present in traffic with no encroachment into the ego lane. For ``\textit{car collision with city guardrails},'' baselines retrieve either collisions that do not involve guardrails, or scenes with guardrails but no collision event, capturing one half of the compositional query while missing the other. {\ours} grounds both the object type and the motion event, returning videos that satisfy the specified conditions in these examples.

\section{Conclusion}
\label{sec:conclusion}

We present {\ours}, a calibrated structured-grounding framework for natural-language retrieval over driving video. The framework rests on a single diagnosis: large language models reliably translate a query into the structural form of an executable rule, but are unreliable at fitting that rule's numerical thresholds to real driving data; this miscalibration is what limits LLM-written rules in practice. {\ours} addresses it by factoring the two competencies: an LLM proposes rule structure, while a learning-to-rank objective recovers the numerical thresholds under distant supervision from captions on an auxiliary in-domain corpus, organizing the calibrated rules into a reusable library that a confidence-gated policy reuses or adapts at query time. Across three driving benchmarks, {\ours} consistently improves over the strongest embedding-based and structured baselines. Several limitations remain: the rule schema lacks explicit temporal sequencing and scoped negation, retrieval depends on upstream detection and tracking, transfer of a calibrated library across perception stacks is untested, and caption errors introduce noise into calibration labels, so thresholds for events with few positives may be unreliable. Despite these limitations, these components together turn LLM-written rules from structurally plausible but numerically miscalibrated proposals into a calibrated, interpretable retrieval primitive grounded in the empirical signal distribution of an unlabeled in-domain corpus.

\clearpage
\begin{ack}
This work was supported by NSF grants CMMI-2326309 and CNS-2312395. The authors declare no additional competing interests.
\end{ack}

\bibliographystyle{unsrt}
\bibliography{ref}

\clearpage
\appendix


\section{Experimental Setup}
\label{sec:settings}

\subsection{Perception Pipeline}
\label{sec:perception_pipeline}

We adopt the perception pipeline of iFinder~\cite{Yao2025iFinder} with a single substitution: the video captioner used at library-construction time is replaced by Qwen2.5-VL-7B~\cite{bai2025qwen25vl}. The pipeline produces per-frame structured representations consumed by the symbolic-rule evaluation in \Cref{sec:dsl-formal}; each record contains a set of tracked objects with object class, 3D distance from the ego vehicle, lane assignment, lateral offset, and additional context fields. The modules and the pretrained model used for each are:
\begin{itemize}
    \item \textbf{Camera intrinsics and lens-distortion estimation.} GeoCalib~\cite{veicht2024geocalib} predicts focal length, principal point, and radial/tangential distortion coefficients per video.
    \item \textbf{Scene understanding and per-object attributes (image VLM).} InternVL~\cite{chen2024internvl} extracts global scene context (weather, road structure, time of day) and per-object attributes such as vehicle color and traffic-light state.
    \item \textbf{Video captioning (video VLM, library-construction only).} Qwen2.5-VL-7B~\cite{bai2025qwen25vl} produces the per-video event description used to seed library construction. This module is invoked only when building the rule library and is not used at retrieval time.
    \item \textbf{Ego-vehicle camera-pose estimation.} DROID-SLAM~\cite{teed2021droidslam} provides per-frame camera translations, from which ego heading-rate and speed are derived to populate the ego turn/motion state.
    \item \textbf{2D object detection.} OWL-V2~\cite{minderer2023owlv2} provides open-vocabulary, class-conditioned bounding boxes.
    \item \textbf{Multi-object tracking.} ByteTrack~\cite{zhang2022bytetrack} associates detections across frames into per-object tracks.
    \item \textbf{Lane detection.} OMR~\cite{jin2024omr} predicts per-frame lane-marking polylines, which are converted to per-object lane assignments by mapping each object's bottom-midpoint into the lane partition.
    \item \textbf{Metric depth.} Metric3D~\cite{hu2024metric3dv2} predicts a metric depth map per frame.
    \item \textbf{Foreground segmentation for distance-to-ego.} SAM~\cite{kirillov2023segmentanything} produces a binary foreground mask within each detected bounding box; the per-object distance is the mean masked depth.
    \item \textbf{3D detection for object orientation.} CenterTrack~\cite{zhou2020centertrack} provides 3D bounding boxes whose yaw angles are matched to the 2D detections via Hungarian assignment.
\end{itemize}

\subsection{Rule Library Construction}
\label{sec:dsl-library}

We build the library from 1{,}147 deduplicated Nexar training videos~\cite{moura2025nexardashcamcollisionprediction}, disjoint from all evaluation benchmarks. Motion events are extracted from captions generated by Qwen2.5-VL-7B~\cite{bai2025qwen25vl} using a predefined inventory of 21 categories; a video may belong to multiple categories. For each category, videos whose captions mention the event form the positive set, and all remaining videos form the negative set.
We calibrate one symbolic rule per category to maximize the AP of the ranking induced by $\sdsl$ over the auxiliary corpus (\Cref{sec:l2r-definitions}), yielding 21 calibrated rules. \Cref{tab:nexar-events} reports the category counts: hard braking and swerving appear in 1{,}068 and 1{,}019 videos, respectively.

\subsection{Rule Scoring: Formal Definitions}
\label{sec:dsl-formal}

Given a symbolic rule $\dslrule$, the motion relevance score $\sdsl(\video, \dslrule)$ is computed in three steps; symbol conventions follow \Cref{sec:method}.

\paragraph{Hard gates.}
With $G$ hard-gate constraints $\{(\phi_g, \mathtt{op}_g, \theta_g)\}_{g=1}^{G}$, where $\phi_g$ is gate $g$'s target field (\eg, \texttt{distance}, \texttt{loc\_x}) and each $\theta_g$ is a threshold calibrated against the auxiliary set, the hard-gate score is
\begin{equation}
    \shard(\video, \dslrule) = \prod_{g=1}^{G} \mathbf{1}\!\left[\, \exists o,\ \exists t:\ \mathtt{op}_g(\phi_g(o, t), \theta_g) \,\right],
    \label{eq:dsl_hard}
\end{equation}
where $\phi_g(o, t)$ is the value of $\phi_g$ for object $o$ at frame $t$. A video is gated to $\shard = 0$ if any single precondition is unsatisfied.

\paragraph{Soft scores.}
For each soft constraint $c$ with target field $\phi_c$, expected magnitude $\delta_c$, and window length $\Delta T_c$, the per-object score for object $o$ is the maximum within-window change in $\phi_c$, normalized by $\delta_c$:
\begin{equation}
    s_c^{(o)} = \min\!\left(1,\ \frac{\max_{0 < t_2 - t_1 \le \Delta T_c} |\phi_c(o, t_2) - \phi_c(o, t_1)|}{\delta_c}\right).
    \label{eq:dsl_soft}
\end{equation}
When constraint $c$ specifies a signed \texttt{direction} (\eg, \texttt{negative} for an approaching object), the absolute value is replaced by the corresponding one-sided difference (\eg, $\max(0,\ \phi_c(o, t_1) - \phi_c(o, t_2))$); \cref{eq:dsl_soft} shows the unsigned default. The aggregate per-object soft score is the weighted average $\strack^{(o)} = \big(\sum_c w_c\, s_c^{(o)}\big) / \big(\sum_c w_c\big) \in [0, 1]$, with weights $w_c$ calibrated jointly with the other numerical constraints.

\paragraph{Entity aggregation and final score.}
The entity selector aggregates per-object scores to a video-level score:
\begin{equation}
    \sentity(\video, \dslrule) =
    \begin{cases}
        \max_{i} \strack^{(i)} & (\textit{any}\text{-mode}) \\[4pt]
        \tfrac{1}{2}\!\left( \max_{i} \strack^{(i)} + \overline{\strack} \right) & (\textit{consensus}\text{-mode})
    \end{cases},
    \label{eq:dsl_entity}
\end{equation}
where $\overline{\strack}$ is the mean across candidate objects. The final motion relevance score is the product $\sdsl(\video, \dslrule) = \shard(\video, \dslrule) \cdot \sentity(\video, \dslrule)$. Hard gates and soft scores are aggregated independently at the video level, so a video can satisfy hard gates via one object and accrue soft scores via another; this decoupling is intentional, with hard gates acting as a global admissibility filter and soft scores capturing graded evidence anywhere in the video.

\begin{table}[t]
\centering
\caption{Extracted driving event categories in the Nexar training set (1{,}147 dashcam videos after de-duplication and canonicalization).}
\label{tab:nexar-events}
\begin{tabular}{lr}
\toprule
\textbf{Event} & \textbf{\# Videos} \\
\midrule
Hard Brake            & 1{,}068 \\
Swerving              & 1{,}019 \\
Pedestrian Crossing   &   370   \\
Near Miss             &   199   \\
Cut-In                &   170   \\
Cut-Out               &   166   \\
Traffic Disruption    &   151   \\
Lane Change           &    96   \\
Object Approach       &    78   \\
Collision             &    62   \\
Emergency Vehicle     &    31   \\
Evasive Maneuver      &    20   \\
Sudden Speed Change   &    18   \\
Lateral Movement      &    10   \\
Sudden Stop           &     9   \\
Loss of Control       &     8   \\
Rollover              &     8   \\
Dynamic Object Crossing &   8   \\
Object Departure      &     3   \\
Reverse               &     1   \\
Sharp Turn            &     1   \\
\midrule
\textbf{Total}        & \textbf{1{,}147} \\
\bottomrule
\end{tabular}
\end{table}

\subsection{L2R Calibration: Precise Definitions}
\label{sec:l2r-definitions}

\paragraph{Selection objective.}
At iteration $\itr$ of the calibration loop, candidate rule $\bdslrule^{(\itr)}$ is scored by the average precision (AP) of the ranking $\ranked$ that $\sdsl$ induces on $\vpos \cup \vneg$:
\begin{equation}
    \mathrm{AP}(\bdslrule^{(\itr)}, \vpos, \vneg) = \frac{1}{|\vpos|} \sum_{k:\, \bvideo_{\ranked_k} \in \vpos} \frac{|\{j \le k : \bvideo_{\ranked_j} \in \vpos\}|}{k},
    \label{eq:ap-suppl}
\end{equation}
where $j$ is the inner running index over $\{1, \ldots, k\}$. AP is the only quantity used to compare proposals; the per-field summary statistics provided to the LLM at proposal time are conditioning inputs only and do not appear in this objective.

\paragraph{Best-so-far feedback.}
We denote the best-scoring rule among the first $\itr$ proposals by
\begin{equation}
    \bdslrule^{(\itr)*} := \arg\max_{\itr' \le \itr} \mathrm{AP}(\bdslrule^{(\itr')}, \vpos, \vneg).
    \label{eq:bestrule-suppl}
\end{equation}
When $\mathrm{AP}(\bdslrule^{(\itr)}, \vpos, \vneg) < \tap$, the LLM is fed the previous AP together with $\bdslrule^{(\itr-1)*}$ as an anchor and asked to propose a revision; this repeats up to $\nrev=20$ iterations.

\paragraph{Calibrated library.}
After calibration, the best assignment per event is recorded as
\begin{equation}
    \library = \{(\beventi, \bdslrule^{(\nrev)*})\}_{\eventi=1}^{\librarysize},
    \label{eq:library-suppl}
\end{equation}
and used at retrieval time according to the confidence-aware policy in \Cref{sec:library}.

\paragraph{Search space and proposal conditioning.}

For each event, the proposal loop searches all four classes of numerical constraints jointly: hard-gate thresholds $\theta_g$, soft-score expected magnitudes $\delta_c$, soft-score weights $w_c$, and window lengths $\Delta T_c$. To keep proposals on the right scale, each LLM proposal step is conditioned on per-field summary statistics (mean, standard deviation, and quantiles) computed from the auxiliary set's per-frame structured representations; these statistics anchor initial guesses but do not enter the AP objective.

\subsection{Multi-Source Fusion: Formal Definition}
\label{sec:fusion-formal}

Per-source rankings from the symbolic-rule path, dense embedding, and sparse lexical retrieval are merged via query-conditioned weighted reciprocal rank fusion~\cite{10.1145/1571941.1572114}:
\begin{equation}
    \sfus(\video) = \sum_{s\, \in\, \{\mathrm{rule},\, \mathrm{dense},\, \mathrm{bm25}\}} \frac{w_s(\query)}{k_{\mathrm{RRF}} + \mathrm{rank}_s(\video)},
    \quad k_{\mathrm{RRF}} = 60,
    \label{eq:fusion}
\end{equation}
where $\mathrm{rank}_s(\video) \in \{1, 2, \ldots\}$ is the position of $\video$ in source $s$'s ranking, $w_s(\query) \in \mathbb{R}_{\ge 0}$ is the query-conditioned weight produced by the fusion router (\Cref{sec:prompt-router}), and $k_{\mathrm{RRF}}$ is the standard RRF smoothing constant~\cite{10.1145/1571941.1572114}. The final ranking is obtained by sorting the candidate pool by $\sfus$ in descending order; a post-fusion reranker is then applied over the top-$\npool$ items (\Cref{sec:exp_settings}).

\subsection{DrivingDojo Event Annotation}
\label{sec:drivingdojo-annotations}

We annotate 583 videos from DrivingDojo~\cite{wang2024drivingdojodatasetadvancinginteractive} with structured event labels and release these annotations at \url{https://github.com/nec-labs-ma/strive-d-data-annotation}.
Each annotation specifies the event type, the object class (ego or surrounding vehicle), and the direction of motion where applicable.
Of the 583 annotated videos, 236 are ground-truth positives for at least one query; the remaining videos serve as non-relevant candidates in the retrieval pool.
The benchmark comprises 41 distinct queries, with each query having between 1 and 138 relevant videos (a video may be relevant to multiple queries).

As shown in \cref{tab:drivingdojo-events}, the queries fall into four event types: \textit{Cut-In}, \textit{Cut-Out}, \textit{Lane Change}, and \textit{Turn}.
Cut-In and Cut-Out queries are further stratified by the surrounding vehicle's object class (car, motorcycle, two-wheeled vehicle, bicycle, truck, and bus), plus a class-agnostic variant; each class-specific query also includes left and right directional sub-queries.
Lane Change and Turn queries involve the ego vehicle and likewise include directional variants.
Each query is expressed as a free-form natural-language description specifying the event type, object class, and direction where applicable.

\paragraph{Annotation protocol.}
Each video is annotated with its event type. For events involving a surrounding vehicle, the annotation also records the vehicle's object class, color (white, grey, black, red, brown, or other), and direction relative to the ego vehicle (left, right, or front). Cut-in labels distinguish entry from the left or right, and cut-out labels distinguish departure to the left or right. Lane-change and turn labels specify left or right motion of the ego vehicle. Two annotators labeled every video, and two additional annotators checked the annotations for quality.

\begin{table}[t]
\centering
\caption{Query categories in the DrivingDojo~\cite{wang2024drivingdojodatasetadvancinginteractive} annotation release.
Cut-In/Cut-Out are stratified by object class (car, motorcycle, two-wheeled, bicycle, truck, bus) plus an class-agnostic variant, each with any/left/right direction variants.
Video count range is across object-class queries; 236 unique videos total.}
\label{tab:drivingdojo-events}
\setlength{\tabcolsep}{8pt}
\begin{tabular}{lrr}
\toprule
\textbf{Event} & \textbf{Queries} & \textbf{\# Videos} \\
\midrule
Cut-In (surrounding vehicle)  & 18 & 2--138 \\
Cut-Out (surrounding vehicle) & 17 & 1--80  \\
Lane Change (ego)             &  3 & 47--98 \\
Turn (ego)                    &  3 & 38--92 \\
\midrule
\textbf{Total / unique videos} & \textbf{41} & \textbf{236} \\
\bottomrule
\end{tabular}
\end{table}

\section{Additional Experiments}
\label{sec:additional_exps}

\subsection{Effect of Candidate Pool Size $K'$}
\label{sec:pool_size}

\begin{figure}[h]
    \centering
    \begin{subfigure}[b]{0.48\linewidth}
        \includegraphics[width=\linewidth]{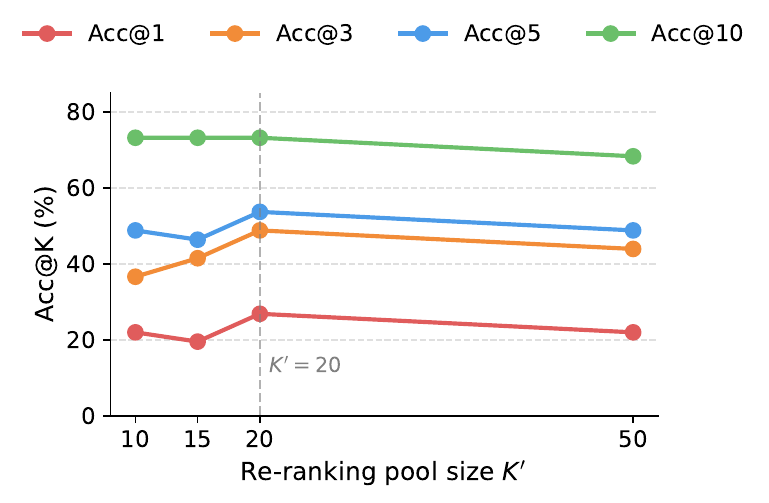}
        \caption{{\ours} (Ours)}
    \end{subfigure}
    \hfill
    \begin{subfigure}[b]{0.48\linewidth}
        \includegraphics[width=\linewidth]{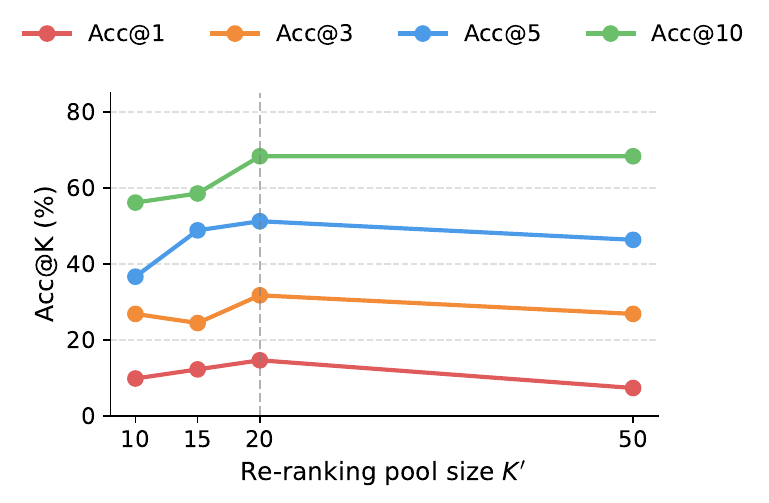}
        \caption{Qwen3-VL-8B}
    \end{subfigure}
    \\[0.5em]
    \begin{subfigure}[b]{0.48\linewidth}
        \includegraphics[width=\linewidth]{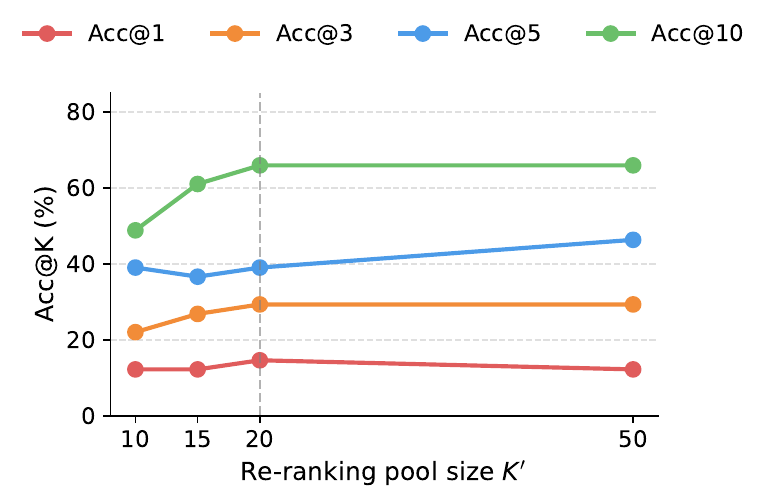}
        \caption{SigLIP2}
    \end{subfigure}
    \hfill
    \begin{subfigure}[b]{0.48\linewidth}
        \includegraphics[width=\linewidth]{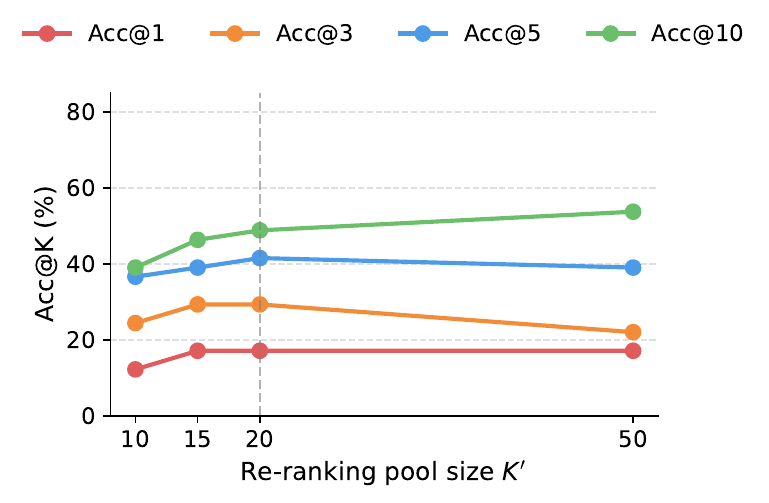}
        \caption{VideoRAG}
    \end{subfigure}
    \caption{Sensitivity to re-ranking pool size $K'$ on DrivingDojo, evaluated for {\ours} and reranker-equipped baselines. Every method either peaks at or is at the plateau by $K'=20$, so we adopt this value uniformly throughout the main paper.}
    \label{fig:pool_size}
\end{figure}

We study how the candidate pool size $K'$ affects retrieval performance for {\ours} and each reranker-equipped baseline (\cref{fig:pool_size}).
All evaluated methods attain strong, near-peak performance at $K'=20$, with further increases yielding only marginal change.
The Acc@$k$ values reported for the reranker-equipped baselines in \cref{tab:main_results} are lifted from this sweep at $K'=20$ for fair comparison; we therefore fix $K'=20$ throughout the main paper.

\subsection{Structured Grounding Choice: Symbolic Rules vs.\ NSVS-TL within Our Pipeline}
\label{sec:dd_struct_choice}

The main-paper comparison in \Cref{tab:main_results} shows that {\ours} outperforms NSVS-TL~\cite{choi2024towards}, but does not isolate \emph{which part} of our framework is responsible for that gap: the calibrated symbolic rules, the multi-source fusion, the reranker, or some combination. To answer this, we compare three structured-retrieval variants on DrivingDojo~\cite{wang2024drivingdojodatasetadvancinginteractive}: NSVS-TL as a standalone retriever in its as-published configuration; NSVS-TL substituted into our pipeline (replacing the symbolic-rule component while keeping dense retrieval, sparse retrieval, the query-conditioned router, and the reranker identical); and full {\ours}.

\begin{table}[!t]
  \centering
  \small
  \setlength{\tabcolsep}{4pt}
  \caption{\textbf{Structured grounding choice (DrivingDojo, \%).} Replacing our calibrated symbolic rules with NSVS-TL's temporal-logic grounding within an otherwise identical fusion pipeline (dense + sparse + query-conditioned router + reranker) loses $14.6$~pp Acc@1 and $4.7$~pp mAP. The precision gain is attributable to the calibrated symbolic-rule component rather than to generic structured grounding.}
  \label{tab:nsvs_compare}
  \begin{tabular}{lcccccc}
    \toprule
    Variant & MRR$\uparrow$ & mAP$\uparrow$ & Acc@1 & Acc@3 & Acc@5 & Acc@10 \\
    \midrule
    NSVS-TL standalone~\cite{choi2024towards} & 31.8 & 4.9 & 22.0 & 34.2 & 43.9 & 53.7 \\
    NSVS-TL within our pipeline & 28.7 & 3.9 & 12.2 & 34.2 & 53.7 & 70.7 \\
    \midrule
    Full {\ours} & \textbf{38.7} & \textbf{8.6} & \textbf{26.8} & \textbf{48.8} & \textbf{53.7} & \textbf{73.2} \\
    \bottomrule
  \end{tabular}
\end{table}

\Cref{tab:nsvs_compare} shows that full {\ours} outperforms NSVS+pipeline by $14.6$~pp Acc@1, $4.7$~pp mAP, and $10.0$~pp MRR despite identical fusion machinery. The gap concentrates at top-of-list precision; broad-recall metrics (Acc@5, Acc@10) are closer because dense and sparse retrieval establish a recall floor independent of the symbolic choice. NSVS+pipeline also drops below NSVS-TL standalone at Acc@1: NSVS-TL emits binary verdicts that our continuous query-conditioned weighted fusion cannot leverage as effectively as graded scores, whereas our symbolic rules produce graded scores by design. This isolates the contribution of the calibrated symbolic rules beyond generic structured grounding: the gain is attributable to the rules' calibrated graded scoring, not to the surrounding fusion machinery.

Beyond accuracy, online inference cost separates the two paradigms decisively: NSVS-TL averages $4{,}868.05$\,s per query because each query requires running its full temporal-logic verification over the corpus, whereas {\ours} averages $3.10$\,s per query, a $\sim$$1{,}500\times$ reduction (\Cref{sec:efficiency}). The cross-dataset NSVS-TL numbers in \Cref{tab:main_results} were obtained at this prohibitive per-query cost; {\ours} reaches the reported accuracy without it.

\subsection{Per-stage Latency Breakdown}
\label{sec:latency_breakdown}

\begin{table}[h]
  \centering
  \small
  \setlength{\tabcolsep}{8pt}
  \caption{\textbf{{\ours} per-stage latency breakdown (mean per query, DrivingDojo).} The two LLM stages, rule synthesis (run only when no library rule matches the query) and multi-source fusion, together account for $\sim$$87\%$ of the total $3{,}096.66$\,ms ($3.10$\,s); the three retrieval branches (sparse, dense, rule matching) cost $\sim$$384$\,ms combined. Both LLM stages are amenable to caching across queries that share an event category.}
  \label{tab:latency_breakdown}
  \begin{tabular}{lr}
    \toprule
    Stage & Latency~(ms) \\
    \midrule
    Sparse retrieval (BM25~\cite{robertson2009probabilistic})   & $1.27$ \\
    Dense retrieval (Qwen3-VL-Embedding-8B~\cite{qwen3vlembedding}) & $128.54$ \\
    Rule matching                                       & $254.16$ \\
    Rule synthesis                                      & $1{,}987.43$ \\
    Multi-source fusion                                 & $725.26$ \\
    \midrule
    \textbf{Total}                                      & $3{,}096.66$ \\
    \bottomrule
  \end{tabular}
\end{table}

\Cref{tab:latency_breakdown} reports the per-stage online latency breakdown of {\ours} on DrivingDojo~\cite{wang2024drivingdojodatasetadvancinginteractive}, complementing the cross-method comparison in \Cref{tab:latency} of the main paper.
The two LLM stages, rule synthesis (invoked only on the exemplar-conditioned fallback path, when no library rule matches the query) and multi-source fusion, together account for $\sim$$87\%$ of the total per-query latency.
Both stages are amenable to caching across queries that share an event category, which our timings do not exploit; the remaining stages (rule matching against per-frame records, dense retrieval, BM25~\cite{robertson2009probabilistic} sparse retrieval) are inexpensive.

\subsection{Variability across Runs}
\label{sec:error-analysis}

Several components of our pipeline are stochastic, including VLM-based
video captioning (\Cref{sec:perception_pipeline}), event extraction
from captions, and the LLM proposer in the calibration loop
(\Cref{sec:l2r-definitions}). To assess the cumulative variability
across these sources, we re-ran library construction and retrieval
five times with independent samples, holding
all other settings fixed, and report mean $\pm$ one standard deviation
(1-$\sigma$) across the five runs in \Cref{tab:variance}. At top-$10$,
the candidate set typically saturates the relevant items, so the
ranking is largely insensitive to small rule perturbations; the
relative ordering against the strongest baseline is preserved in
every run.

\begin{table}[h]
  \centering
  \small
  \setlength{\tabcolsep}{8pt}
  \caption{\textbf{Variability of {\ours} across 5 independent runs.} Mean $\pm$ one standard deviation (1-$\sigma$); variability captures the cumulative effect of all stochastic components in the pipeline (VLM captioning, event extraction, LLM proposer). Standard deviations stay below $1.3$ points for Acc@$\{1,3,5\}$ and below $0.3$ for Acc@$10$ on all three benchmarks.}
  \label{tab:variance}
  \begin{tabular}{lcccc}
    \toprule
    Dataset & Acc@1 & Acc@3 & Acc@5 & Acc@10 \\
    \midrule
    MMAU~\cite{Fang_2024_CVPR}                                       & $26.5 \pm 1.1$ & $53.1 \pm 1.1$ & $59.2 \pm 1.1$ & $65.3 \pm 0.0$ \\
    DrivingDojo~\cite{wang2024drivingdojodatasetadvancinginteractive} & $26.8 \pm 1.3$ & $48.8 \pm 1.1$ & $53.7 \pm 1.1$ & $73.2 \pm 0.0$ \\
    CarCrash~\cite{BaoMM2020}                                        & $28.9 \pm 1.2$ & $60.0 \pm 1.1$ & $73.3 \pm 0.9$ & $75.6 \pm 0.3$ \\
    \bottomrule
  \end{tabular}
\end{table}

\subsection{Proposer Comparison: Protocol}
\label{sec:proposer_protocol}
We compare calibration proposers with each rule's entity selector and the fields used by its hard gates and soft scores fixed. All calibrated variants optimize AP on the Nexar auxiliary set~\cite{moura2025nexardashcamcollisionprediction} (\Cref{eq:ap-suppl}), with a budget of 20 candidate parameter assignments per rule. The assignment with the highest auxiliary AP is retained.
\begin{itemize}
  \item \textbf{Uncalibrated baseline:} thresholds are fixed at $40\%$ of each field's range, with no search.
  \item \textbf{Random search:} sample 20 assignments, drawing thresholds uniformly from their allowed ranges and sampling weights, directions, and gate settings at random.
  \item \textbf{Bayesian search:} evaluate 10 random assignments, followed by 10 proposals concentrated around the best assignment found so far.
  \item \textbf{Grid search:} evaluate the 20 combinations of minimum change at $\{15,30,45,60\}\%$ of each field's range and distance gate in $\{20,30,40,50,60\}$\,m.
  \item \textbf{LLM proposer:} generate assignments conditioned on per-field statistics and best-so-far feedback, following \Cref{sec:l2r-definitions}.
\end{itemize}
Random and Bayesian search are each evaluated over three runs; \Cref{tab:proposer} reports mean $\pm$ standard deviation. Each resulting library is evaluated with the same matcher, adapter, fusion, and reranker.

\subsection{Perception Robustness: Protocol and Full Results}
\label{sec:robustness_full}
We perturb one module at a time in the precomputed perception records and rerun DrivingDojo~\cite{wang2024drivingdojodatasetadvancinginteractive} retrieval with five random seeds. The calibrated library, remaining fields, and dense and sparse paths are fixed.

\paragraph{Independent perturbations (\Cref{tab:robust_indep}).}
We randomly select $P\%$ of detected-object entries for removal, tracks for fragmentation, or vehicles for depth perturbation. Selected tracks are split at random points, with new identities assigned to their tails. Depth perturbation multiplies the selected vehicles' distances by factors drawn from $\mathcal{N}(1,(P/100)^2)$.

\paragraph{Temporally correlated perturbations (\Cref{tab:robust_real}).}
False detections are added at $P\%$ of the original detection count and persist for 1--5 frames. Their classes are sampled from the clip, and their boxes and depths are copied from real objects with small noise. Consecutive misses remove $P\%$ of each object's detections in one contiguous block with a random start. Track interruptions remove $P\%$ of tracks for $k\in\{10,20\}$ frames, after which they resume under new identities.

\begin{table}[h]
  \centering
  \small
  \setlength{\tabcolsep}{8pt}
  \caption{\textbf{Independent perception corruption} (DrivingDojo Acc@10, \%; mean $\pm$ std over 5 seeds; uncorrupted $=73.2$).}
  \label{tab:robust_indep}
  \begin{tabular}{lccc}
    \toprule
    Corrupted module & $P=0$ & $P=25$ & $P=50$ \\
    \midrule
    Depth     & $73.2$ & $72.7 \pm 3.2$ & $72.2 \pm 3.3$ \\
    Detection & $73.2$ & $71.2 \pm 3.2$ & $70.2 \pm 2.0$ \\
    Tracking  & $73.2$ & $70.7 \pm 1.7$ & $69.8 \pm 1.3$ \\
    \bottomrule
  \end{tabular}
\end{table}

\begin{table}[h]
  \centering
  \small
  \setlength{\tabcolsep}{8pt}
  \caption{\textbf{Temporally correlated perception corruption} (DrivingDojo Acc@10, \%; mean $\pm$ std over 5 seeds; uncorrupted $=73.2$).}
  \label{tab:robust_real}
  \begin{tabular}{lccc}
    \toprule
    Corruption & $P=25$ & $P=50$ & $P=75$ \\
    \midrule
    False detections                    & $68.3 \pm 3.5$ & $72.4 \pm 3.0$ & $69.1 \pm 3.0$ \\
    Consecutive missed detections       & $67.5 \pm 2.3$ & $64.2 \pm 2.3$ & $62.6 \pm 7.5$ \\
    Track re-establishment ($k=10$)     & $71.7 \pm 1.3$ & $69.8 \pm 1.3$ & $66.7 \pm 1.2$ \\
    Track re-establishment ($k=20$)     & $70.7 \pm 3.5$ & $64.2 \pm 4.1$ & $63.4 \pm 3.1$ \\
    \bottomrule
  \end{tabular}
\end{table}


\subsection{Choice of Rule-Generator LLM and Reranker}
\label{sec:llm_choice}
To assess sensitivity to model choice, \Cref{tab:abla_model} compares rule-generator LLMs and Qwen3-VL rerankers~\cite{qwen3vlembedding} on DrivingDojo~\cite{wang2024drivingdojodatasetadvancinginteractive}. Acc@10 remains $73.2\%$ across the tested variants, while Acc@1 varies with the generator: $26.8\%$ for GPT-4.1~\cite{openai2025gpt41}, $22.0\%$ for GPT-4o~\cite{openai2024gpt4ocard}, and $17.1\%$ for GPT-4o-mini~\cite{openai2024gpt4omini}. Replacing the default 2B reranker with the 8B model lowers Acc@1, MRR, and mAP, with Acc@5 and Acc@10 unchanged. For the embedding baselines, \Cref{tab:abla_reranker} reports the effect of adding Qwen3-VL-Reranker, which raises Acc@1 for both model sizes on all three datasets.

\begin{table}[!t]
 \centering
 \small
 \setlength{\tabcolsep}{4pt}

 \caption{\textbf{Rule-generator and reranker comparison (DrivingDojo, \%).} We vary the rule-generator LLM or the reranker while keeping the other components fixed.}
 \label{tab:abla_model}
  \scalebox{0.85}{%
 \begin{tabular}{lcccccc}
 \toprule
 \textbf{Method} & \textbf{MRR} & \textbf{mAP} & \textbf{Acc@1} & \textbf{Acc@3} & \textbf{Acc@5} & \textbf{Acc@10} \\
 \midrule
 \multicolumn{7}{c}{\textit{Rule Generator Choice}} \\
 \midrule
 Ours (w/ GPT-4o-mini) & 30.08 & 3.47 & 17.07 & 36.59 & 51.22 & 73.17 \\
 Ours (default, GPT-4.1) & \textbf{38.7} & \textbf{8.6} & \textbf{26.8} & \textbf{48.8} & \textbf{53.7} & \textbf{73.2} \\
 Ours (w/ GPT-4o)      & 32.01 & 3.47 & 21.95 & 36.59 & 48.78 & 73.17 \\
\midrule
 \multicolumn{7}{c}{\textit{Reranker}} \\
 \midrule
 Ours (default, 2B Reranker) & \textbf{38.7} & \textbf{8.6} & \textbf{26.8} & \textbf{48.8} & \textbf{53.7} & \textbf{73.2} \\
 Ours (w/ 8B Reranker) & 33.86 & 5.29 & 17.07 & 39.02 & 53.66 & 73.17 \\
 \bottomrule
 \end{tabular}}
\end{table}

\begin{table}[!t]
 \centering
 \small
 \setlength{\tabcolsep}{4pt}
 \caption{\textbf{Effect of Qwen3-VL-Reranker on the embedding baselines (\%).} For each dataset, we report Qwen3-VL-2B and Qwen3-VL-8B~\cite{qwen3vlembedding} with and without their corresponding Qwen3-VL-Reranker. The +Reranker rows use $\npool=20$ and match the main results (\cref{tab:main_results}).}
 \label{tab:abla_reranker}
 \begin{tabular}{lcccccc}
 \toprule
 Method & MRR($\uparrow$) & mAP($\uparrow$) & Acc@1 & Acc@3 & Acc@5 & Acc@10 \\
 \hline

 \multicolumn{7}{c}{Dataset: \textit{DrivingDojo}\cite{wang2024drivingdojodatasetadvancinginteractive}} \\
 \midrule
 QWen3-vl-2B & 10.0 & 2.3 & 4.3 & 14.6 & 24.4 & 46.3 \\
 QWen3-vl-2B + Reranker & 12.2 & 1.4 & 4.9 & 26.8 & 46.3 & 65.9 \\
 QWen3-vl-8B & 15.8 & 1.2 & 4.9 & 19.5 & 31.7 & 56.1 \\
 QWen3-vl-8B + Reranker & 20.1 & 2.2 & 14.6 & 31.7 & 51.2 & 68.3 \\
 \midrule

 \multicolumn{7}{c}{Dataset: \textit{MM-AU}~\cite{Fang_2024_CVPR}} \\
 \midrule
 QWen3-vl-2B & 17.8 & 4.0 & 8.2 & 20.4 & 30.6 & 46.9 \\
 QWen3-vl-2B + Reranker & 26.9 & 6.0 & 18.4 & 30.6 & 42.9 & 49.0 \\
 QWen3-vl-8B & 34.6 & 8.6 & 18.4 & 32.7 & 44.9 & 53.1 \\
 QWen3-vl-8B + Reranker & 32.7 & 6.3 & 20.4 & 38.8 & 55.1 & 61.2 \\
 \midrule

 \multicolumn{7}{c}{Dataset: \textit{CarCrashDataset}~\cite{BaoMM2020}} \\
 \midrule
 QWen3-vl-2B & 31.9 & 24.8 & 17.8 & 37.8 & 48.9 & 64.4 \\
 QWen3-vl-2B + Reranker & 38.3 & 32.1 & 26.7 & 44.4 & 55.6 & 64.4 \\
 QWen3-vl-8B & 34.9 & 27.5 & 20.0 & 44.4 & 55.6 & 66.7 \\
 QWen3-vl-8B + Reranker & 40.1 & 33.5 & 28.9 & 46.7 & 62.2 & 66.7 \\
 \bottomrule
 \end{tabular}
\end{table}

\section{Additional Qualitative Results}

\begin{figure}[t]
  \centering
  \setlength{\tabcolsep}{2pt}
  \renewcommand{\arraystretch}{0.5}
  \newcommand{\labelours}{\rotatebox{90}{\small\textcolor{ForestGreen}{\textbf{Ours\hspace{6pt}}}}}
  \newcommand{\labelsiglip}{\rotatebox{90}{\small SigLIP2\hspace{4pt}}}
  \newcommand{\labelqwen}{\rotatebox{90}{\small Qwen3-8B\hspace{0pt}}}
  \newcommand{\iw}{0.44\textwidth}

  \begin{tabular}{c cc}
    & \small\textit{truck skidded in the rain} & \small\textit{car collided with a rock at a city intersection} \\[3pt]
    \labelours &
    \includegraphics[width=\iw]{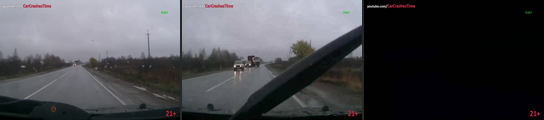} &
    \includegraphics[width=\iw]{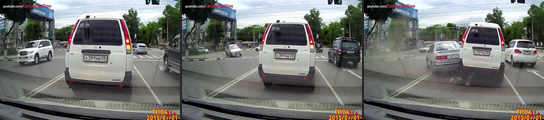} \\[3pt]
    \labelsiglip &
    \includegraphics[width=\iw]{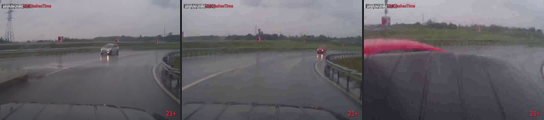} &
    \includegraphics[width=\iw]{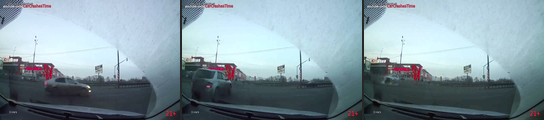} \\[3pt]
    \labelqwen &
    \includegraphics[width=\iw]{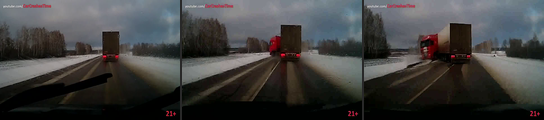} &
    \includegraphics[width=\iw]{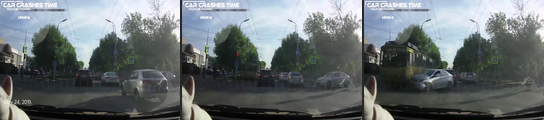} \\
  \end{tabular}

  \vspace{6pt}

  \begin{tabular}{c cc}
    & \small\textit{car drove into the field} & \small\textit{truck fell into the ditch} \\[3pt]
    \labelours &
    \includegraphics[width=\iw]{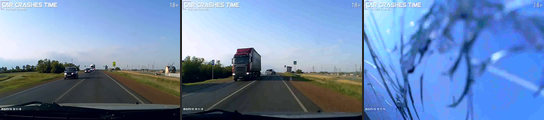} &
    \includegraphics[width=\iw]{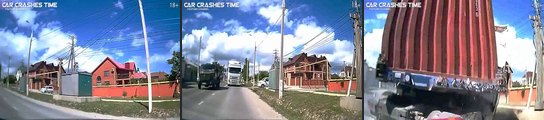} \\[3pt]
    \labelsiglip &
    \includegraphics[width=\iw]{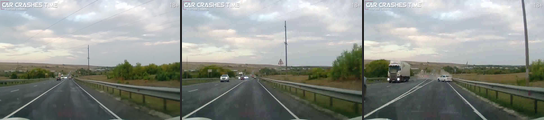} &
    \includegraphics[width=\iw]{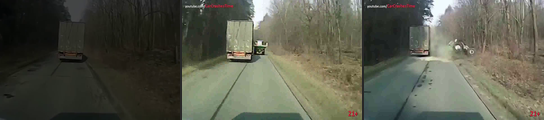} \\[3pt]
    \labelqwen &
    \includegraphics[width=\iw]{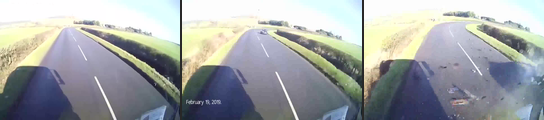} &
    \includegraphics[width=\iw]{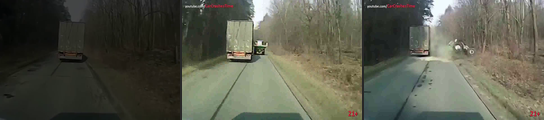} \\
  \end{tabular}

  \caption{\textbf{Qualitative retrieval comparison.} Sampled frames from the top-1 retrieved video per method for four complex natural-language queries on DrivingDojo~\cite{wang2024drivingdojodatasetadvancinginteractive}; our method consistently surfaces semantically relevant videos while vision-language baselines (SigLIP2~\cite{tschannen2025siglip}, Qwen3-VL~\cite{qwen3vlembedding}) struggle with queries involving environmental conditions or rare event types.}
  \label{fig:suppl_qual}
\end{figure}

\cref{fig:suppl_qual} presents additional qualitative comparisons on four complex queries from the DrivingDojo~\cite{wang2024drivingdojodatasetadvancinginteractive} dataset.
Our method retrieves videos matching the query, while vision-language baselines (SigLIP2~\cite{tschannen2025siglip}, Qwen3-VL~\cite{qwen3vlembedding}) return less relevant results on these examples.

\begin{figure}[t]
  \centering
  \setlength{\tabcolsep}{3pt}
  \renewcommand{\arraystretch}{0.6}
  \newcommand{\iwd}{0.48\textwidth}

  \begin{tabular}{c cc}
    & \small\textit{car skids in snow} & \small\textit{motorcycle cutin} \\[3pt]
    \rotatebox{90}{\small CarCrash\hspace{6pt}} &
    \includegraphics[width=\iwd]{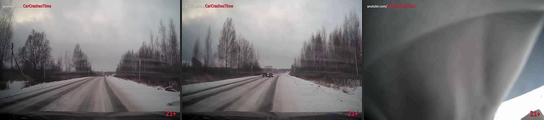} &
    \includegraphics[width=\iwd]{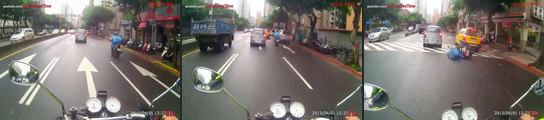} \\[6pt]
    \rotatebox{90}{\small DrivingDojo\hspace{2pt}} &
    \includegraphics[width=\iwd]{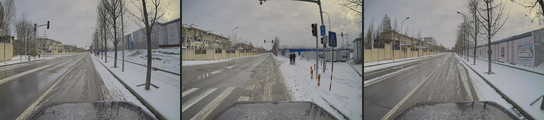} &
    \includegraphics[width=\iwd]{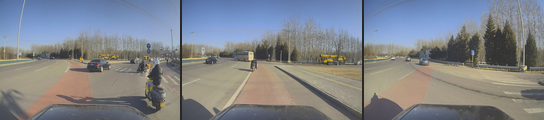} \\[6pt]
    \rotatebox{90}{\small MM-AU\hspace{8pt}} &
    \includegraphics[width=\iwd]{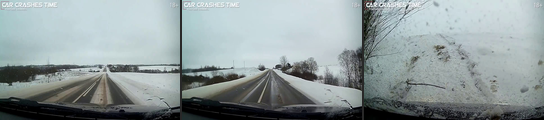} &
    \includegraphics[width=\iwd]{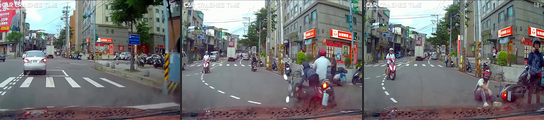} \\
  \end{tabular}

  \caption{\textbf{Cross-dataset retrieval results beyond ground-truth annotations.}
  For each query and dataset (CarCrash~\cite{BaoMM2020}, DrivingDojo~\cite{wang2024drivingdojodatasetadvancinginteractive}, MM-AU~\cite{Fang_2024_CVPR}), we show sampled frames from the top-1 retrieved video.
  The retrieved videos are not present in the ground-truth annotations, yet appear visually consistent with the query.}
  \label{fig:cross_datasets}
\end{figure}

\paragraph{Cross-dataset retrievals beyond ground-truth annotations.}
\cref{fig:cross_datasets} presents examples in which our method retrieves videos that fall outside the ground-truth annotations, yet appear visually consistent with the query.
For instance, the query \textit{``car skids in snow''} retrieves snowy road-incident videos from CarCrash~\cite{BaoMM2020}, DrivingDojo~\cite{wang2024drivingdojodatasetadvancinginteractive}, and MM-AU~\cite{Fang_2024_CVPR}; the query \textit{``motorcycle cutin''} retrieves motorcycles encroaching on the ego lane across all three datasets.
These examples also illustrate that the same calibrated rule library can be applied across dataset boundaries without dataset-specific adaptation.

\section{LLM Prompts}
\label{sec:prompts}

This section documents the seven LLM prompt templates used by {\ours}, grouped by pipeline stage. Library-construction prompts (\Cref{sec:prompt-captioner,sec:prompt-event,sec:prompt-l2r-init,sec:prompt-l2r-revision}) run once on the auxiliary set to caption videos, extract candidate events, and calibrate the rules' numerical constraints. Retrieval-time prompts (\Cref{sec:prompt-matcher,sec:prompt-translator,sec:prompt-router}) run per query to match against the calibrated library, synthesize an adapted rule when no library entry matches, and produce per-source weights that combine the symbolic, dense, and sparse retrieval signals through weighted reciprocal rank fusion~\cite{10.1145/1571941.1572114}.

\subsection{Video Captioner}
\label{sec:prompt-captioner}

The captioner runs at library construction only, on each video in the auxiliary set, to produce the natural-language description $\bvcap$ that downstream event extraction (\Cref{sec:prompt-event}) parses for weak-supervision labels. The prompt targets kinematic content (spacing, movement, signals, scene adjustments) rather than generic scene description; even with this prompting, captions remain a noisy motion-event signal, which motivates the AP-averaged calibration objective (\Cref{sec:l2r-definitions}). \Cref{lst:p-cap} shows the prompt.

\begin{lstlisting}[caption={Video captioner system prompt: chronological motion-targeted description used at library construction.}, label={lst:p-cap}]
You are an autonomous driving expert, specializing in recognizing traffic scenes.

You receive a series of traffic images captured from the perspective of the ego car. Your task is to analyze the video and provide a detailed, chronological description of the scene.

For the scene description, for each significant event or change, describe:
- Changes in Spacing: Note how the distance between vehicles and stable objects changes over time. Indicate if vehicles get closer to or farther from these objects.
- Movement Dynamics: Explain the movement of each vehicle, including their direction, any noticeable adjustments in their position relative to other vehicles and stationary elements, and whether the vehicle is approaching or stopping.
- Vehicle Signals: Pay attention to whether vehicles have their blinkers or lights on, and note their status in your description.
- Scene Adjustments: Describe any significant changes in the arrangement of vehicles and objects and their impact on the overall scene.

Ensure that the description captures the sequence of events as they occur in the video, detailing vehicle behavior and scene dynamics without referring to individual frames.
\end{lstlisting}

\subsection{Event Extraction}
\label{sec:prompt-event}

For each auxiliary video, the extractor parses the captioner's description into a list of (event\_type, subject) candidates with a short evidence quote. After consolidation across the auxiliary set, the deduplicated list forms the event index $\sevent = \{\bevent_1, \ldots, \bevent_\librarysize\}$ used to construct the calibrated library $\library$. Per-event positive and negative video sets $\vpos, \vneg$ are derived by matching extracted events against each caption: a video is positive for $\beventi$ if its extracted events contain $\beventi$, and negative otherwise. \Cref{lst:p-event} shows the prompt.

\begin{lstlisting}[caption={Event extraction system prompt: emits a JSON array of (event\_type, subject, confidence, evidence) tuples per caption.}, label={lst:p-event}]
You are an expert in driving video analysis and vehicle kinematics.

Given a driving video summary, extract ALL notable kinematic events -- i.e.,
events that involve measurable physical motion that could be captured by
sensors (speed, distance, lateral position, trajectory).

INCLUDE events involving:
- Sudden longitudinal speed changes (hard braking, rapid acceleration, emergency stop)
- Lateral movements (cut-in, cut-out, lane change, swerving)
- Object approach / departure (vehicle closing in, near miss, overtaking)
- Dynamic object crossing (pedestrian, cyclist, animal crossing the road)
- Emergency vehicle passing (siren, flashing lights with motion)
- Construction zone entry / slowing

DO NOT include:
- Static scene descriptions with no motion (parked car, road type, weather alone)
- Object presence without motion relevance (car visible in mirror but not interacting)
- Ambiguous non-kinematic observations

Output a JSON array (can be empty []).  Each element must have:
{
  "event_type": "<snake_case label, e.g. hard_brake / cutin / pedestrian_crossing>",
  "subject":    "ego" | "object" | "both",
  "confidence": "high" | "medium" | "low",
  "evidence":   "<short exact quote or paraphrase from the summary>"
}

Return ONLY the JSON array, no explanation.
\end{lstlisting}

\subsection{L2R Proposer: Initial Proposal}
\label{sec:prompt-l2r-init}

For each event $\beventi$ extracted in \Cref{sec:prompt-event}, the L2R loop's first iteration calls the proposer with the event's natural-language label and per-field summary statistics from the auxiliary set, producing a complete candidate rule (entity selector, hard gates, soft scores, agg\_mode). Subsequent iterations use the revision prompt (\Cref{sec:prompt-l2r-revision}). Threshold values in the few-shot examples are illustrative structures only; the proposer is expected to substitute dataset-specific values from the field-statistics block. \Cref{lst:p-init} shows the prompt.

\begin{lstlisting}[caption={L2R proposer initial-proposal system prompt: emits a complete symbolic rule conditioned on the event label and per-field summary statistics.}, label={lst:p-init}]
You are a driving video retrieval expert. Translate a natural-language query about driving behavior into a structured JSON symbolic rule that can deterministically score every video in a dataset.

## Dataset context
Dataset: {DATASET_NAME} (dashcam videos, ego-centric, 10 fps, typical length 10-30 s).
Task: rank videos by relevance to the query; the rule will be scored against tracked-object time-series extracted from JSON annotations.

## Available fields (per tracked object, per frame)

| Field      | Source            | Unit   | Notes |
|------------|-------------------|--------|-------|
| distance   | 2D monocular depth| m      | Integer-quantized |
| loc_z      | 3D detector       | m      | Forward depth. Only for tracked objects. |
| loc_x      | 3D detector       | m      | Lateral offset: positive = right of ego, negative = left. |
| lane_rel   | VLM text label    | lanes  | Relative lane vs ego: 0=same, +1=one right, -1=one left. |
| bbox_area  | 2D detector       | px^2   | Bounding box of objects. |


### agg_mode="any" (object events: cut-in, approach, hard-brake)
Your threshold MUST be calibrated so that only rare individual tracks satisfy it:
- A threshold at the P50 means ~50% of all tracks will satisfy it -- far too permissive.
- For a query matching ~20% of videos: aim for P75-P85 (~15-25% of tracks exceed).
- For a rare event (<10%): target P90-P95.
- Use 'pct_exceeding' rows to read off: "if I set min_delta=X, what fraction of tracks will pass?"

Example: distance P50-change-in-4s = 12 m and 57% of tracks exceed 10 m ->
min_delta=10 is too permissive for a rare proximity event. Use 30 m (P80) instead.


## Entity spec
{
  "role": "object",          // "object" = a detected other vehicle/pedestrian
  "class_filter": ["car"],   // null = any class; valid: car, truck, bus, pedestrian, motorcycle, bicycle, barrier, traffic_cone
  "selection": "any",        // "any" = evaluate all matching tracks
                             // "closest" = only the nearest object by distance
  "agg_mode": "any"          // HOW to aggregate scores across multiple tracks:
                             // "any"       = max over tracks (default)
                             //              Use when ONE specific object causes the event
                             //              (cut-in, hard-brake, pedestrian crossing)
                             // "consensus" = mean over tracks when min_fraction pass the gate
                             //              Use for EGO-MOTION events (lane change, turn):
                             //              when ego changes lane, ALL nearby objects shift
                             //              in the same direction simultaneously
}

NOTE: This dataset does NOT have an explicit ego-lane-index field.
To detect ego maneuvers (lane change, turn), use surrounding objects' relative motion as proxy:
  - Ego changes lane RIGHT -> nearby objects shift LEFT (lane_rel decreases, loc_x decreases)
  - Ego turns LEFT -> nearby objects' loc_x shifts RIGHT
  - For ego-motion queries, ALWAYS set "agg_mode": "consensus" -- the pattern should appear
    across most surrounding vehicles, not just a lucky single track.

## Hard gates (binary, video scores 0 if none of its tracks pass)
{ "field": "distance", "op": "lt", "value": 40.0 }
Operators: "lt" (<), "le" (<=), "gt" (>), "ge" (>=).
Use gates only when there is a clear physical reason (e.g., object must be within 40 m for a near-miss).

## Soft Scores (continuous in [0, 1])
soft_scores: weighted evidence terms that produce a graded relevance in [0, 1] per video (unlike hard_gates which are binary pass/fail).

Each entry tests one kinematic signal on one field and scores smoothly:
  signal == threshold -> 0.5;  signal == 2x threshold -> 1.0;  wrong direction -> 0.

Per-track scores are aggregated by entity.agg_mode:
  "any"       -> max  (one object suffices: cut-in, brake, pedestrian crossing)
  "consensus" -> mean over tracks, gated by min_fraction (ego-motion: lane change, turn)

The final video score is the weight-weighted average of all soft_scores.
Use weight to express relative importance; use min_fraction (not min_delta) to discriminate ego-motion from noise.


## Output format
Return ONLY a valid JSON object -- no markdown, no comments, no text outside the JSON:

{
  "event": "short_snake_case_event_name",
  "confidence": 0.85,
  "entity": {
    "role": "object",
    "class_filter": ["car", "truck"],
    "selection": "any",
    "agg_mode": "any"
  },
  "hard_gates": [...],
  "soft_scores": [...],
  "scoring_notes": "One sentence explaining the rule's logic, threshold choices, direction, and agg_mode."
}

## Few-shot examples
Note: threshold values in these examples are illustrative STRUCTURES only.
Always substitute thresholds from the {FIELD_STATS} section above for the actual dataset.

### Query: "ego car change lane to the right"
{
  "event": "ego_lane_change_right",
  "confidence": 0.80,
  "entity": {
    "role": "object",
    "class_filter": ["car", "truck", "bus"],
    "selection": "any",
    "agg_mode": "consensus"
  },
  "hard_gates": [],
  "soft_scores": [
    {
      "type": "monotonic_trend",
      "field": "loc_x",
      "direction": "decreasing",
      "min_duration_seconds": 2.0,
      "min_fraction": 0.20,
      "weight": 1.0
    }
  ],
  "scoring_notes": "Ego moves right -> ALL nearby objects shift LEFT (loc_x decreasing). min_fraction=0.20 requires 20% of tracks to show a 2+ s sustained decrease simultaneously. In noise ~25% of tracks show this in some window; but a real lane change shifts ALL tracks consistently, so a much higher fraction agree -> better-ranked videos. NO temporal_change -- window-based change is dominated by noise for ego-motion."
}

### Query: "object approaching ego"
{
  "event": "approach",
  "confidence": 0.90,
  "entity": {
    "role": "object",
    "class_filter": null,
    "selection": "any",
    "agg_mode": "any"
  },
  "hard_gates": [],
  "soft_scores": [
    {
      "type": "monotonic_trend",
      "field": "distance",
      "direction": "decreasing",
      "min_duration_seconds": 3.0,
      "weight": 1.0
    },
    {
      "type": "temporal_change",
      "field": "distance",
      "min_delta": 30.0,
      "window_seconds": 5.0,
      "direction": "negative",
      "weight": 0.8
    }
  ],
  "scoring_notes": "One specific object approaches. direction=negative on temporal_change counts only distance reductions (not random fluctuation). agg_mode=any -- only the approaching object needs to show this signal. Threshold at >=P85 of distance change."
}

Now translate the following query. Return ONLY the JSON object.
\end{lstlisting}

\subsection{L2R Proposer: Revision}
\label{sec:prompt-l2r-revision}

When a candidate assignment fails self-consistency checks on the auxiliary set, the proposer is fed back the previous AP, a percentile summary of the score distribution, and a list of issue tags, and asked to revise. \Cref{lst:p-rev} shows the message.

\begin{lstlisting}[caption={L2R proposer revision user message: feeds back the previous AP, score-distribution percentiles, and a list of issue tags, with the best-so-far assignment as the anchor for the next proposal.}, label={lst:p-rev}]
The previous rule for query '{query}' achieved AP={ap_t} on the
auxiliary set (target >= {tap}).
Best-so-far AP: {ap_best}; use the best-so-far rule as the anchor for
this revision.

Issue tags detected:
{issues}

Score distribution across {N} auxiliary videos:
  P10={p10}  P25={p25}  P50={p50}  P75={p75}  P90={p90}  P95={p95}
  nonzero (>0.05): {nonzero_ratio} of videos

Please revise the rule with these adjustments:
{hints}

Keep the same event type. Return ONLY a valid JSON object.
\end{lstlisting}

\subsection{Library Matcher}
\label{sec:prompt-matcher}

The library matcher maps a natural-language query to one of 21 predefined event categories and returns a scalar confidence score.
Queries with confidence $\ge \tauh$ ($\tauh = 0.6$) are routed directly to the matched library template; queries with confidence $< \tauh$ trigger the rule translator, which generates an adapted rule using the highest-confidence library template as a few-shot exemplar. \Cref{lst:p1-system} shows the prompt.

\begin{lstlisting}[caption={Library matcher system prompt: event categories, disambiguation rules, and JSON output format.}, label={lst:p1-system}]
You are a driving event classifier for a video retrieval system.

Given a natural-language query about a driving incident, identify which of the
following event categories best describes the PRIMARY event. If no category
fits well, return "none".

Event categories:
- hard brake: The EGO vehicle decelerates suddenly/abruptly.
- swerving: The EGO vehicle weaves or makes erratic lateral movements
  (all surroundings shift together -- camera effect).
  NOT for a surrounding vehicle that moves sideways; use "lateral movement".
- lane change: The EGO or another vehicle smoothly changes lanes.
- evasive maneuver: The EGO driver steers/brakes sharply to avoid an obstacle.
- sudden speed change: The EGO vehicle abruptly accelerates or decelerates.
- sudden stop: The EGO vehicle stops abruptly from speed.
- loss of control: A vehicle loses control: spins out, drifts. If the query
  adds a condition the kinematic rule cannot encode (snow/rain/ice/"for no
  apparent reason"), cap confidence at 0.55.
- rollover: A vehicle rolls over, tips onto its side, or flips.
- reverse: A vehicle is reversing/backing up.
- sharp turn: A vehicle makes a sharp or aggressive turn.
- cutin: A SURROUNDING vehicle cuts in front of the ego car
  (lane-change encroachment). NOT a general collision.
- cutout: A SURROUNDING vehicle ahead moves out of the ego lane.
- lateral movement: A SURROUNDING vehicle (not the ego) moves noticeably sideways.
- object approach: A surrounding object rapidly approaches the ego.
- object departure: A surrounding object moves away from the scene.
- traffic disruption: A vehicle blocks or disrupts traffic flow.
- collision: Direct physical impact: vehicle-vehicle or vehicle-object crash.
- near miss: Close call without actual contact.
- pedestrian crossing: A pedestrian crosses in front of the ego.
- emergency vehicle: Ambulance, fire truck, or police is present.
- dynamic object crossing: Animal, debris, or non-vehicle object crosses the road.

Disambiguation rules:
- "swerving" = EGO weaves; all surroundings shift together (camera).
  "lateral movement" = a SURROUNDING vehicle moves sideways. Not the same.
- "cutin"/"cutout" = lane-change encroachments only -- not crashes.
- "collision" = any direct physical impact.
- "rollover" = rolls, tips, flips, falls on its side.
- Condition-modifier rule: weather/road conditions (snow, rain, ice) or vague
  causes ("for no apparent reason") set confidence <= 0.55, so the system
  generates a fresh adapted rule instead of reusing the library template.
- Return "none" only when the query describes no driving incident at all.

Respond with valid JSON only -- no extra text:
{"event": "<category_or_none>", "confidence": <0.0-1.0>, "reason": "<one sentence>"}
\end{lstlisting}

\subsection{Rule Adapter}
\label{sec:prompt-translator}

When the library matcher confidence falls below $\tauh$, the rule adapter produces a query-specific rule by adjusting the highest-confidence library template under dataset field statistics, treating the template as a few-shot exemplar. The system prompt below (Listing~\ref{lst:p2-system}) specifies the rule schema, constraint types, and few-shot examples; the user message (Listing~\ref{lst:p3-user}) prepends the matched template alongside the new query.

\begin{lstlisting}[caption={Rule adapter system prompt: task definition, field statistics, rule schema, constraint types, and few-shot examples.}, label={lst:p2-system}]
You are a driving video retrieval expert. Translate a natural-language query
about driving behavior into a structured JSON symbolic rule that can
deterministically score every video in a dataset.

Dataset context
Dataset: [DATASET_NAME] (dashcam videos, ego-centric, 10 fps, 10-30s).
Task: rank videos by relevance to the query; the rule will be scored against
tracked-object time-series extracted from JSON annotations.

Available fields (per tracked object, per frame):
- distance   : forward distance (m), 2D depth
- loc_z      : forward depth (m), 3D detection
- loc_x      : lateral offset (m), 3D detection; +=right of ego
- lane_rel   : lane offset (lanes), VLM; 0=same, +1=right, -1=left
- bbox_area  : bounding-box area (px^2), 2D detection

CRITICAL: Data-driven field statistics for this dataset
[Automatically generated from the target dataset. Contains, for each field
(distance, loc_x, loc_z, lane_rel) and each time window (2s, 4s, 8s):
percentile distribution of max|Delta value| across all tracked objects, plus
the fraction of tracks exceeding representative thresholds in a 4s window.
Example snippet (DrivingDojo):]

How to use the statistics to set thresholds:
- A threshold at P50 means ~50% of tracks satisfy it -- too permissive for rare events.
- For a query matching ~20% of videos: aim for P75-P85 (~15-25% of tracks exceed).
- For a rare event (<10%): target P90-P95.

Entity spec:
{
  "role": "object",
  "class_filter": ["car"],   // null = any; valid: car, truck, bus,
                             //   pedestrian, motorcycle, bicycle,
                             //   barrier, traffic_cone
  "selection": "any",        // "any" = all matching tracks
  "agg_mode": "any"          // "any"       = max over tracks (ONE object causes event)
                             // "consensus" = mean when min_fraction pass (EGO-MOTION)
}

Constraint types:

temporal_change -- max field change within any sliding window:
{
  "type": "temporal_change",
  "field": "distance",
  "min_delta": 30.0,          // set from field stats (P80-P95)
  "window_seconds": 4.0,
  "direction": null,          // null=absolute | "positive" | "negative"
  "min_fraction": null,       // only for consensus: fraction of tracks >= 0.5
  "weight": 1.0
}

Hard gates (binary; video scores 0 if no track passes):
{ "field": "distance", "op": "lt", "value": 40.0 }
Operators: "lt" (<), "le" (<=), "gt" (>), "ge" (>=).
Use gates only when there is a clear physical reason.

Output format -- return ONLY a valid JSON object, no markdown:
{
  "event": "short_snake_case_event_name",
  "confidence": 0.85,
  "entity": { "role": "object", "class_filter": [...],
              "selection": "any", "agg_mode": "any" },
  "hard_gates": [...],
  "soft_scores": [...],
  "scoring_notes": "One sentence explaining the rule's logic, thresholds, and agg_mode."
}

Few-shot examples (thresholds are dataset-specific; substitute from FIELD_STATS):

Query: "other car cutin"
{
  "event": "cutin",
  "confidence": 0.90,
  "entity": { "role": "object", "class_filter": ["car","truck"],
              "selection": "any", "agg_mode": "any" },
  "hard_gates": [{"field": "distance", "op": "lt", "value": 40.0}],
  "soft_scores": [
    { "type": "temporal_change", "field": "loc_x", "min_delta": 14.0,
      "window_seconds": 4.0, "direction": null, "weight": 1.0 },
    { "type": "temporal_change", "field": "distance", "min_delta": 20.0,
      "window_seconds": 4.0, "direction": "negative", "weight": 0.7 }
  ],
  "scoring_notes": "Lateral shift >= P90 loc_x combined with approach; agg_mode=any."
}

Query: "ego car change lane to the right"
{
  "event": "ego_lane_change_right",
  "confidence": 0.80,
  "entity": { "role": "object", "class_filter": ["car","truck","bus"],
              "selection": "any", "agg_mode": "consensus" },
  "hard_gates": [],
  "soft_scores": [
    { "type": "temporal_change", "field": "loc_x", "min_delta": 3.5,
      "window_seconds": 4.0, "direction": "negative",
      "min_fraction": 0.20, "weight": 1.0 }
  ],
  "scoring_notes": "Ego right => objects shift left (negative loc_x); min_fraction=0.20 for consensus agreement; agg_mode=consensus."
}

Query: "object approaching ego"
{
  "event": "approach",
  "confidence": 0.90,
  "entity": { "role": "object", "class_filter": null,
              "selection": "any", "agg_mode": "any" },
  "hard_gates": [],
  "soft_scores": [
    { "type": "temporal_change", "field": "distance", "min_delta": 30.0,
      "window_seconds": 5.0, "direction": "negative", "weight": 1.0 }
  ],
  "scoring_notes": "direction=negative counts reductions only; agg_mode=any."
}

Now translate the following query. Return ONLY the JSON object.
\end{lstlisting}

\begin{lstlisting}[caption={Rule adapter user message: injects the closest library template as an initialization seed alongside the new query.}, label={lst:p3-user}]
A similar event from the rule library is "[matched_event_key]"
(keyword similarity=[sim]).
Use this rule as a starting point and adapt it to the new query.
Adjust thresholds to match the target dataset's field statistics,
and change entity/class_filter as needed.

Reference rule:
[JSON of the closest library template]

Query: <natural-language query>
\end{lstlisting}

\subsection{Fusion Router}
\label{sec:prompt-router}

The fusion router outputs a per-source weight for three retrieval backends: a dense vision-language embedding (Backend A), the symbolic-rule path (Backend B), and a sparse lexical retriever over per-video summaries (Backend C). The prompt frames weight assignment as a perception-alignment decision: which backend's input modality contains the discriminator that separates matching from non-matching videos for this query? The LLM identifies a primary backend whose modality carries the discriminator and optionally adds a supplementary weight to Backend C when the query has surface tokens (a color, a model, a place name) likely to appear in summary text. Three geometric facts about the dashcam setup (ego-motion produces global pixel shifts; surrounding-object motion produces local pixel changes whose visibility depends on end-of-motion location; summary text rarely contains technical event names) anchor the decision in physical priors rather than free-form reasoning. \Cref{lst:p-router} shows the prompt.

\begin{lstlisting}[caption={Fusion router prompt: outputs per-source weights for dense (Backend A), symbolic rule (Backend B), and sparse lexical (Backend C) retrieval.}, label={lst:p-router}]
For this query, assign per-source weights for fusion across three retrieval
backends.  Each backend produces its own ranking; the weights you output
determine how much each ranking contributes to the final ordering.
The three backends, identified by neutral labels A, B, C, are each defined
by what they perceive and nothing more:

  - Backend A : perceives pixels.  Ranks each candidate video by the
                 visual similarity between its dashcam frames and the
                 implied content of the query, as a human would judge it
                 by looking at the frames.  Returns degenerate ranks when
                 matching and non-matching frames look the same to a
                 human.

  - Backend B : has no access to pixels.  Perceives, for each video, a
                 per-frame numeric record of detected objects (each
                 object's position relative to the ego camera, distance,
                 over time).  Ranks each candidate by whether this
                 numeric record matches a pattern derived from the query.
                 Returns informative ranks precisely when the visual
                 frames are NOT enough to discriminate matching from
                 non-matching cases -- when the discriminator is in the
                 numeric record only.

  - Backend C : perceives a single natural-language summary string per
                 video.  Ranks candidates by token overlap with the
                 query.  Useful only when the discriminating words in
                 the query also literally appear in the matching summaries
                 (and not in non-matching ones).

Geometric facts about the dashcam (apply these directly):

  Fact 1.  The camera is rigidly mounted on the ego vehicle.  When EGO
  moves (rotates or translates), every pixel in the frame shifts coherently
  -- lane markings sweep, the horizon tilts, every off-road landmark slides.
  This is a global change in pixels that Backend A picks up immediately.

  Fact 2.  When the ego is straight-and-steady and another object moves,
  only the pixels covering THAT object change.  Whether Backend A picks up
  this local pixel change depends on where the object ends up:
    - End position in the central viewport (large, close, ahead): the
      object dominates the frame, matching frames look distinct from
      non-matching ones, Backend A discriminates.
    - End position off-centre (alongside ego in a parallel lane, no longer
      interacting with ego): matching frames look the same as ordinary
      multi-lane driving, Backend A's signal is degenerate; only Backend B
      (which sees the numeric record) can discriminate.

  Fact 3.  Dashcam summary text is generic ("a car drives along ...").
  Technical event names ("cutin", "cutout", "lane change") rarely appear
  in summaries even when the event occurs, so Backend C is informative only
  when the query has a non-technical token (a color, a model, a place).

Reasoning protocol:

Step 1 -- Decode the physical motion in ONE short sentence:
  (a) the moving subject (ego, or a specific other object class)
  (b) the direction of motion in ego's reference frame
  (c) where the subject is at the END of the motion

Step 2 -- Identify the PRIMARY backend whose perception aligns with the
discriminator for this query.  Apply Facts 1 and 2 to decide which
backend's input modality contains the discriminator that separates
matching from non-matching videos.  The primary backend should receive
the highest weight.

Step 3 -- Decide whether to add a SUPPLEMENTARY weight to Backend C.
Inspect the query's surface tokens: if the query contains non-technical
tokens (a color, a model, a place name) likely to appear literally in
summary text (Fact 3), assign Backend C a small supplementary weight;
otherwise leave it at zero.

Step 4 -- Output JSON of the form
    {"dense": <float>, "rule": <float>, "bm25": <float>}
where the value at each label is the weight you have assigned to that
backend (Backend A <-> "dense", Backend B <-> "rule", Backend C <-> "bm25").

Query: "{query}"

Output Step 1, Step 2 (cite which Fact and why), Step 3 (count surface
tokens and decide whether Backend C should receive supplementary weight),
and on a final line output ONLY the JSON object.
\end{lstlisting}

\section{Limitations and Broader Impact}
\label{sec:limitations_impact}

\subsection{Limitations}
\label{sec:limitations}

Our rule schema lacks explicit temporal sequencing and scoped negation, so queries requiring these operators rely on dense or lexical retrieval, or approximate synthesized rules. Retrieval also depends on upstream detection and tracking, particularly under temporally correlated failures (\Cref{sec:robustness}); transfer of a calibrated library across perception stacks remains untested. VLM caption errors introduce noise into calibration labels, and thresholds for events with few caption-derived positives may be unreliable (\Cref{tab:nexar-events}). Linear scans of per-frame records and up to two LLM calls per query may limit scalability beyond the evaluated corpora (up to $1{,}953$ videos). Finally, the contribution of the LLM fusion router is not isolated, as we have not compared it with uniform or validation-tuned fixed weights.

\subsection{Broader Impact}
\label{sec:broader_impact}

{\ours} targets natural-language retrieval over driving video; plausible positive uses include scenario mining for autonomous-driving development, retrieval of rare or safety-critical events for regulatory and post-incident analysis, and curation of training or evaluation corpora for downstream perception and planning. The same capability could be repurposed for driver surveillance or privacy-eroding video mining, and biases in the upstream perception pipeline (\eg, demographic disparities in detection or tracking) propagate into retrieval results without being corrected by our calibration. To mitigate these risks, we recommend two safeguards: (i) restricting source corpora to data with appropriate consent and provenance, and (ii) auditing the upstream perception pipeline for such biases before deployment. More broadly, our results suggest that numerical calibration is a binding constraint in applying LLM-written rules to physically grounded retrieval, that data-driven calibration closes this gap without sacrificing the interpretability of rule-based scoring, and that this separation of rule construction and numerical calibration should extend to other domains with similar per-frame structured representations, such as surgical, sports, or industrial monitoring. Each application would require a separate assessment of domain-specific risks.


\end{document}